\documentclass[a4paper,journal]{IEEEtran}
\ifCLASSINFOpdf
  \usepackage[pdftex]{graphicx}
  \DeclareGraphicsExtensions{.pdf,.jpeg,.png}
\else
  \usepackage[dvips]{graphicx}
  \DeclareGraphicsExtensions{.eps}
\fi
\usepackage[cmex10]{amsmath}
\usepackage{amssymb}

\usepackage{tikz}
\usepackage{textcomp}
\usepackage{hyperref}
\usepackage{lipsum}

\newcommand\copyrighttext{ 
  \footnotesize \textcopyright 2016 IEEE. Personal use of this material is permitted.
  Permission from IEEE must be obtained for all other uses, in any current or future
  media, including reprinting/republishing this material for advertising or promotional
  purposes, creating new collective works, for resale or redistribution to servers or
  lists, or reuse of any copyrighted component of this work in other works.
  DOI: \href{<http://ieeexplore.ieee.org>}{<DOI No. TIP-16200-2016>}}
\newcommand\copyrightnotice{%
\begin{tikzpicture}[remember picture,overlay]
\node[anchor=south,yshift=10pt] at (current page.south) {\fbox{\parbox{\dimexpr\textwidth-\fboxsep-\fboxrule\relax}{\copyrighttext}}};
\end{tikzpicture}
}

\begin{document}
%
\title{Track Everything: Limiting Prior Knowledge in Online Multi-Object Recognition}
%
%
%



\author{\IEEEauthorblockN{Sebastien C. Wong\IEEEauthorrefmark{1},~\IEEEmembership{Senior Member,~IEEE},
Victor Stamatescu\IEEEauthorrefmark{2},~\IEEEmembership{Member,~IEEE}, 
Adam Gatt\IEEEauthorrefmark{3},~\IEEEmembership{Member,~IEEE},
David Kearney\IEEEauthorrefmark{2},
Ivan Lee\IEEEauthorrefmark{2}~\IEEEmembership{Senior Member,~IEEE} and
Mark D. McDonnell\IEEEauthorrefmark{2},~\IEEEmembership{Senior Member,~IEEE}}
\IEEEauthorblockA{\IEEEauthorrefmark{1} Defence Science and Technology Group, Edinburgh, SA, Australia}
\IEEEauthorblockA{\IEEEauthorrefmark{2} Computational Learning Systems Laboratory, 
  School of Information Technology and Mathematical Sciences,
  University of South Australia, Mawson Lakes, SA, Australia}
\IEEEauthorblockA{\IEEEauthorrefmark{3} Australian Defence Force, Edinburgh, SA, Australia}
\thanks{Manuscript received September 26, 2016, accepted April 9 2017.
 Corresponding authors: S.~C.~Wong (email: sebastien@computer.org), V.~Stamatescu (email: victor.stamatescu@unisa.edu.au)}}

\maketitle
\copyrightnotice

\begin{abstract}
  This paper addresses the problem of online tracking and classification of multiple objects in an image sequence.
  Our proposed solution is to first track all objects in the scene
  without relying on object-specific prior knowledge,
  which in other systems can take the form of hand-crafted features or user-based track initialization.
  We then classify the tracked objects with a fast-learning image classifier that is based on a shallow convolutional neural network architecture
  and demonstrate that object recognition improves when this is combined with object state information from the tracking algorithm.
  We argue that by transferring the use of prior knowledge from the detection and tracking stages to the classification stage
  we can design a robust, general purpose object recognition system
  with the ability to detect and track a variety of object types.
  We describe our biologically inspired implementation, which adaptively learns the shape and motion of tracked objects,
  and apply it to the Neovision2 Tower benchmark data set, which contains multiple object types. 
  An experimental evaluation demonstrates that our approach is competitive
  with state-of-the-art video object recognition systems that do make use of object-specific prior knowledge in detection and tracking,
  while providing additional practical advantages by virtue of its generality.
\end{abstract}

\begin{IEEEkeywords}
object recognition, image classification, visual tracking, multi-object tracking
\end{IEEEkeywords}

%
\IEEEpeerreviewmaketitle

\section{Introduction}
\label{sec:intro}
We report on the design of an automated vision system that can accurately locate and recognize multiple types of objects.
The goal of online object recognition systems is to continuously detect and correctly classify
the objects in a scene as they undergo changes in motion or appearance.
Furthermore, the system should be robust to distracting or occluding clutter.
Our proposed solution to these challenges is an adaptive multiple object tracking (MOT) algorithm
that tracks all objects in the scene and defers any decisions on \emph{what is an object of interest} to a separate classification stage.
Object recognition then involves combining these class predictions, 
with state information given by object tracking.
This approach emulates the separate \emph{what} and \emph{where} processing streams in primate vision~\cite{wilson1993primate},
and allows the tracking process to be performed without any reliance on object-specific prior knowledge.

An important practical consideration in the design of online object recognition systems is the finite amount of labeled and annotated data available for training.
When scarce, this can degrade classification performance due to overfitting and reduce the detection probability of highly tuned object detectors.
Even when larger data sets are available, these may be biased in such a way
that their image statistics do not accurately reflect the data encountered by the system at run time~\cite{torralba2011bias}.
In the case of classifier-based object recognition~\cite{razavian2014cnnfeatures} and detection~\cite{girshick2014cnnfeatures},
the use of \emph{features}, which are higher-level representations of an object than the raw image, can mitigate these problems
by providing a degree of invariance across different data sets.
In the case of tracking and object detection algorithms,
the same set of challenges can be addressed by making the tracker and detector designs less domain-specific.
In our system this is achieved through the use of \emph{adaptive tracking} (e.g.~\cite{comaniciu2003kernel,wong2005act})
and by employing a \emph{track-before-detect}~\cite{baker1998uncertainty} approach that 
delays the requirement for object specific prior knowledge from detection until recognition. 

We note that there exist commercial and security video analysis applications
in which the user may not possess specific knowledge about new, previously unseen objects.
For example, the user may not have access to information on the appearance of a set of target objects,
but may still wish to track these targets in order to accumulate a domain-specific data set.
Moreover, it may be impractical for the user to initialize the system on multiple targets,
especially when more objects are expected to come into view, or are stationary for long periods.
Therefore, in applications where the system requirements are initially not well defined,
a useful first step is for the system to autonomously detect and track all (moving and stationary) objects, 
including those that may, at first, not be considered objects of interest.

Given these aims and real-world requirements,
we present a novel approach to online object recognition centered on the idea of tracking all salient objects in the scene.
We argue that this ``track everything'' approach can be realized by limiting the explicit use of prior knowledge,
and demonstrate that this can be implemented by simultaneously learning both feature and spatial information about each object
and assigning new measurements to system tracks. 
This argument is supported by the following contributions:
\begin{itemize}
\item a novel object shape learning algorithm, the Shape Estimating Filter (SEF),
  and its multi-object counterpart, the Competitive Attentional Correlation Tracker using Shape (CACTuS)~\cite{wong2009shape};
\item the integration of a feature learning (FL) algorithm with a shape learning algorithm~\cite{gatt2010online}; 
\item {CACTuS-FL}: the first algorithm to automatically detect and track multiple objects in a video sequence
  without object-specific prior knowledge~\cite{wong2014modeldrift};
\item an online object recognition system that employs an ensemble of single hidden layer feedforward networks (SLFNs)
  to combine state information from the multi-object tracking algorithm ({CACTuS-FL})
  with the output from an image classifier, the Shallow Convolutional Neural Network (S-CNN).
\end{itemize}

The rest of this paper is organized as follows.
Key recent advances in the areas of multi-object tracking,
image classification and object recognition systems are outlined in \mbox{Section \ref{sec:review}}.
An overview of our system is provided in \mbox{Section \ref{sec:overview}}, 
and this is expanded upon in \mbox{Sections \ref{sec:features}} to \ref{sec:what}.  
We demonstrate and examine the efficacy of our approach using Neovision2 benchmark data in \mbox{Section \ref{sec:evaluation}}.
Finally, \mbox{Section \ref{sec:conclusion}} concludes the paper with a summary of our findings.

\section{Related work}
\label{sec:review}

We review related works in the areas of online multi-object detection and tracking, object recognition, and benchmarks for evaluating such systems.

\subsubsection*{Online detection and tracking}
Recent state-of-the-art online multi-object trackers
(e.g.~\cite{yang2009people,breitenstein2011multiperson,wu2013agreement,bae2014tracklet,jodoin2014urbantracker})
follow the \emph{tracking-by-detection} approach,
where objects of interest are detected independently in each frame and then uniquely associated with system tracks from the previous frame.
The term \emph{online} implies that the underlying algorithm may only use information collected up to the current frame.
The aforementioned examples rely on specialised people detectors, with the exception of {Urban Tracker}~\cite{jodoin2014urbantracker},
which uses background subtraction to detect all types of traffic under the assumption that only moving objects are of interest.  
This assumption of motion can also be used to form tracklets~\cite{arandjelovic2011contextually}, elementary trajectory fragments, 
which can clustered together (usually in an off-line manner) to form complete tracks. 
Although tracking-by-detection algorithms are state-of-the-art,
one limitation stems from noisy or missed detections,
which can lead to incomplete system tracks.
New systems generally aim to mitigate this problem
through more reliable object detector design and/or better data association techniques.
For example, {Breitenstein \emph{et al.}}~\cite{breitenstein2011multiperson} handled occlusions by coupling detection confidence maps
with an association scheme based on online-learned classifiers.
{Bae \& Yoon}~\cite{bae2014tracklet} used tracklet confidence to resolve unreliable detections,
while their data association stage was based on online discriminative appearance learning.
Unlike the aforementioned examples,
our system relies instead on the track-before-detect paradigm~\cite{baker1998uncertainty}, which is less prone to missing weak detections.
Under this approach, the tracking process guides the detection process in order to correlate detections over multiple frames.

\subsubsection*{Recognition}
Our approach to object recognition is motivated by the success of deep learning for image classification tasks (see~\cite{schmidhuber2015review} for a recent review). 
This typically involves training deep (multi-layered) hierarchical models
such as Deep Belief Networks (DBNs)~\cite{hinton2006dbn} and Convolutional Neural Networks (CNNs)~\cite{lecun1998gradient}.
By training complex models with large amounts of data
CNNs have set new image classification benchmarks in recent years
through models such as AlexNet~\cite{krizhevsky2012alexnet}, OverFeat~\cite{sermanet2014overfeat} and VGGNet~\cite{simonyan2014vggnet}.
Rather than relying on such deep architectures, however,
our system performs object recognition using a Shallow CNN~\cite{mcdonnell2015overfeat} that limits learning to a single layer.
It has been shown to achieve competitive results on standard image classification data sets~\cite{mcdonnell2015elm}
while being fast to train (when compared with standard deep learning approaches)
and maintaining low implementation complexity (few tuneable metaparameters).

\subsubsection*{Benchmark data}
The third key ingredient to our system is domain-specific image sequence data
with sufficient object class labeled examples to allow the supervised training of S-CNNs.
As mentioned previously, most public multi-object tracking data sets,
including those collected for the recent MOT Challenge~\cite{leal2015motchallenge}, contain only a single (pedestrian) target class.
This focus on people tracking is highlighted by the latest data release, MOT16~\cite{milan2016mot16},
in which ground truth object classes are grouped into three broad categories:
Target (pedestrian, cyclist, skater), Ambiguous (lying/sitting person, reflection, distractor), Other (car, motorbike, occluder, bicycle).
An image sequence data set that does contain multiple object types has been provided by the DARPA Neovision2~\cite{neovision2} program.
This data set was collected to enable training and evaluation of
Neuromorphic Vision algorithms~\cite{katsuri2014neuromorphicvision,paiton2012procstreams,khosla2014neuromorphic,cao2015spiking},
which are a class of object recognition algorithms motivated by the emergence of bio-inspired vision sensors~\cite{posch2014retinomorphic}
and processing hardware (e.g.~\cite{merolla2014millionspiking}).

\subsubsection*{Prior knowledge}
As previously discussed, in a tracking-by-detection approach ~\cite{yang2009people,breitenstein2011multiperson,wu2013agreement,bae2014tracklet} object specific prior knowledge is embedded into the detector model. 
Another common prior assumption is that only moving objects are of interest, leading to detection through background subtraction~\cite{jodoin2014urbantracker}, or track formation through  tracklets~\cite{arandjelovic2011contextually}.  
These assumptions limit tracking to only a specific set of objects, or only moving objects.
Furthermore, offline trackers not only make use of prior knowledge of objects, but also incorporate knowledge about future frames, and thus can not run on streaming video.
For object recognition using a CNN~\cite{simonyan2014vggnet,mcdonnell2015elm} prior knowledge is strongly embedded into these models through the large training data sets. 
Thus, there is sufficient scope within the literature to investigate an online system design that transfers prior knowledge from detection and tracking into recognition.

\section{Overview}
\label{sec:overview}

\begin{figure}
 \begin{center}
 \includegraphics[ height=9.5cm]{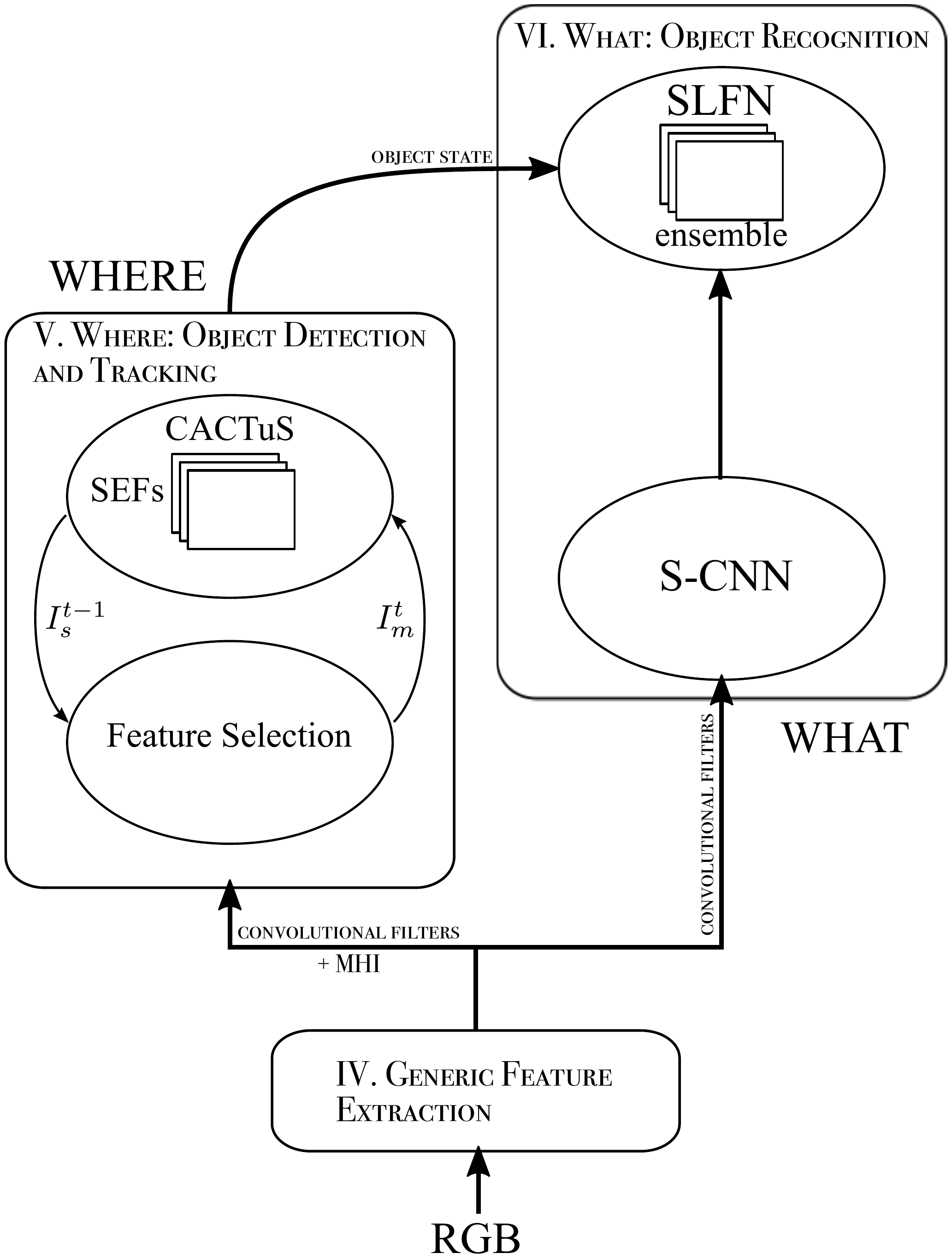}
 \end{center}
 \caption{ \label{fig:system_overview} Overview of our system for online object recognition
   comprising the \emph{where} ({CACTuS-FL})
   and \emph{what} (S-CNN and SLFN ensemble) processing streams.
   The SLFN also combines object state information from the \emph{where} stream.
  }
\end{figure}

This section provides an overview of our online object recognition system, shown in Figure~\ref{fig:system_overview}, as well the notation used in this paper. 

\subsubsection*{Road map}

\mbox{Section \ref{sec:features}} describes the generic feature extraction stage that is used by the \emph{what} and \emph{where} processing streams.
The \emph{where} processing stream (Section \ref{sec:where}) seeks to locate salient objects in the scene
and guide the attention of the \emph{what} processing stream (Section \ref{sec:what}) to these objects.
The \emph{where} stream is handled by the autonomous multi-object tracking algorithm {CACTuS-FL}~\cite{wong2014modeldrift}.
The \emph{what} processing stream relies on a S-CNN architecture~\cite{mcdonnell2015overfeat}
that is followed by an ensemble of SLFNs~\cite{mcdonnell2015elm},
which combines the S-CNN output with object state information from the \emph{where} processing stream.
The S-CNN and individual SLFNs are trained offline
and then deployed in the online classification of image regions (or patches) associated with system tracks.

\subsubsection*{Notation}

Probability mass functions
(PMFs) are denoted by capital letters. The subscripts $p$, $m$,
\& $s$ are used to denote predicted, measured and posterior 
PMFs respectively, while the subscript $0$ denotes a constant prior.
The superscripts $t$ and $t-1$ denote the current and previous time
frames respectively. For brevity, equations that operate only on the current frame
do not include superscript $t$. 
The notation for normalizing across all bins $\boldsymbol{u}$ of a histogram 
to form a PMF is abbreviated to $\frac{1}{\Sigma_{\boldsymbol{u}}}$ to avoid additional indexing variables.

\section{Generic Feature Extraction}
\label{sec:features}

Good features are those which provide a response that discriminates the object(s) of interest and is invariant to changes in the scene. Here we desire a set of common features that are good for both detection and recognition. Furthermore, for our track everything approach every candidate object (including clutter and stationary objects) should be tracked, and is therefore of interest.  

Our tracker, {CACTuS-FL}, can operate on any set of features, including hand-crafted features~\cite{wong2014modeldrift}, however, recent experimental evidence demonstrates that convolutional filters learned by CNNs can produce good features for online visual tracking, enhancing state-of-the-art performance~\cite{danelljan2015conv,ma2015hierconvfeatures}. 
Furthermore, while motion provides a strong visual cue to the presence of salient objects,   
which can form an image feature~\cite{yin2006mhi} or constrain appearance models~\cite{martin2010MOTmotion}, 
this type of cue can not, by itself, detect stationary objects.

For object recognition, the orderless pooling of CNN filter banks can also provide state-of-the-art performance~\cite{cimpoi2015deep},
despite earlier evidence to the contrary~\cite{varma2003texture}.

Thus, we choose a motion history image (MHI) feature~\cite{yin2006mhi}, as moving (as well as stationary) objects are of interest, and a biologically inspired convolutional filter bank~\cite{stamatescu2016filters} that is learned in a generative manner to encapsulate the entire scene.

\subsection{Motion History Image}

The MHI~\cite{yin2006mhi} combines object movement information over an image sub-sequence. To meet the requirement of online tracking we avoid the backward MHI and implement only the forward MHI. This candidate feature is obtained from frame differences between the current image and historical images (through a Markov chain), which highlights the cumulative object motion with a gradient trail that fades away.

\subsection{Convolutional Filters}

\begin{figure}
 \begin{center}
 \includegraphics[trim = 25mm 85mm 25mm 85mm, clip, height=3.8cm]{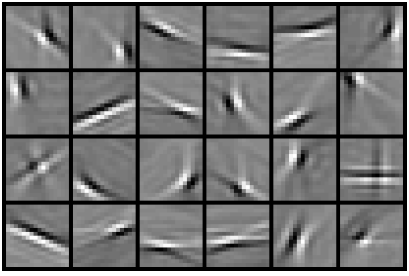}
 \end{center}
 \caption{ \label{fig:crbm_filters} Bank of $24$ generative filters of size $16 \times 16$ pixels learned
  using a Convolutional Restricted Boltzmann Machine (CRBM)~\cite{hlee2009crbm}.
  The unsupervised training was carried out using
  the first frames of Neovision2 Tower training sequences.
  All training image were first converted to greyscale and pre-processed (see main text for details).
  }
\end{figure}

\begin{figure}
  \begin{center}
  \includegraphics[trim = 60mm 123mm 60mm 123mm, clip, height=4.5cm]{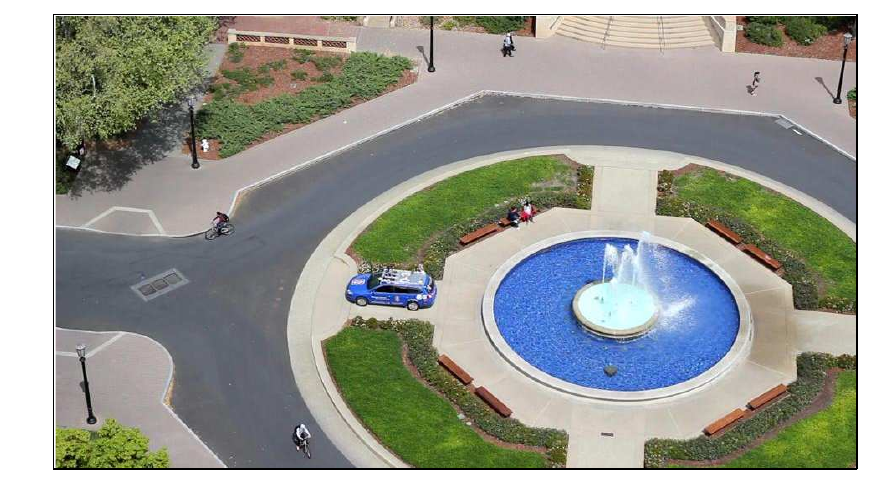}
  \includegraphics[trim = 60mm 123mm 60mm 123mm, clip, height=4.5cm]{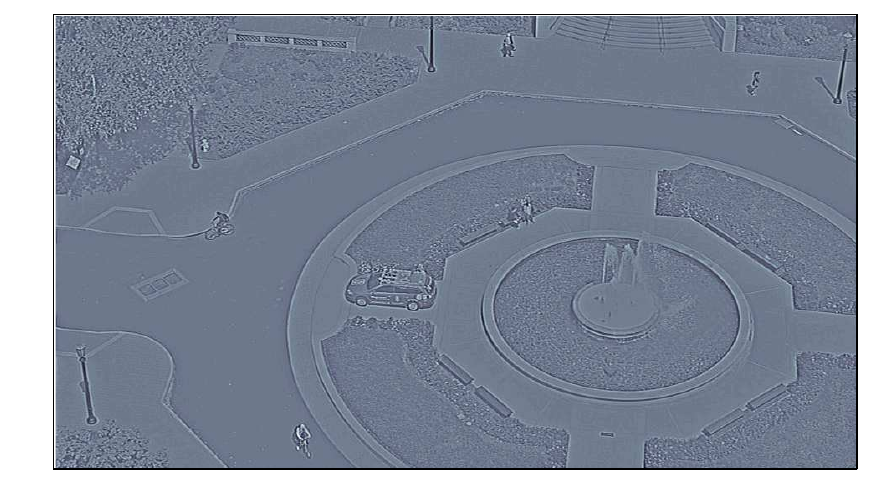}
\end{center}
\caption{ \label{fig:frame61_input}
  Sample RGB (top) and pre-processed (bottom) input image,
  showing frame $61$ from Neovision2~\cite{neovision2} Tower image sequence $001$,
  which was first downsampled to a size of $960\times540$ pixels.
  Image pre-processing involves the application of the whitening function used by Olshausen \& Field~\cite{olshausen1997sparsecoding},
  subtraction of the image mean and normalization of the result by its root mean square (\emph{rms}).}
\end{figure}

The $24$ convolutional filters, shown in Figure~\ref{fig:crbm_filters}, were learned in an unsupervised manner
from the first frames of Neovision2 Tower training image sequences $010-024$
by using a Convolutional Restricted Boltzmann Machine (CRBM)~\cite{hlee2009crbm}.
Each greyscale filter has dimensions of $16 \times 16$ pixels, which was chosen empirically~\cite{stamatescu2016filters}.
In training the generative CRBM model,
RGB input images were first downsampled by a factor of two (to a size of $960\times540$ pixels)
to match the resolution of input images used in the online object recognition system.
The training images were pre-processed by converting to greyscale,
applying the whitening function used by Olshausen \& Field~\cite{olshausen1997sparsecoding},
subtracting the image mean and normalizing the result by its root mean square (\emph{rms}),
as illustrated in Figure~\ref{fig:frame61_input}.
The whitening function applies a combined whitening and low-pass filter with frequency response of the form $fe^{-(f/f_{0})^{4}}$,
where $f_{0}$ is a cutoff frequency of $200$ cycles/image.
During the online application of these filters, each new input image also undergoes these pre-processing steps.

\section{Where: Object Detection and Tracking}
\label{sec:where}

Multiple object tracking algorithms are required to maintain temporally consistent trajectories (state information) for all objects
and to uniquely associate new observations with each trajectory.
An additional requirement in our design is that tracks are able to self-initialize
by automatically converging onto regions of temporally consistent and spatially correlated local saliency.
To this end, we couple the track-before-detect paradigm with an adaptive tracking approach (e.g.~\cite{comaniciu2003kernel,wong2005act}),
so that a state model, which recursively learns both object shape and motion, is able to guide future detections.
The unique identities of multiple objects are preserved by correctly associating multiple sub-trackers with new observations.
This is accomplished by operating these sub-trackers in competition with one another across the scene.

\subsection{Feature Selection}

We first address the problem of autonomous single object detection.
Typical object detectors in visual tracking use application-specific knowledge
such as hard-coding a fixed set of features that describe a particular object or type of object.
By contrast, this paper follows the adaptive method proposed by Collins \emph{et al.}~\cite{collins2005online},
which frames the online selection of a subset of features (from a larger set) as an evolving ``object versus local background'' two-class classification problem.
This \emph{discriminant tracking} approach is analogous to the center-surround mechanisms for attention and saliency
that are found in biological vision~\cite{mahadevan2013saliency} and enable automatic track initiation.

Every candidate feature $n \in {1,...,25}$ (the MHI feature and the $24$ convolutional features from Section~\ref{sec:features})
is used to compute a \emph{feature map} $Z_{n}^{t}\left(\boldsymbol{i}\right)$,
which is a representation of the image at frame $t$ in terms of the feature response at each pixel position $\boldsymbol{i}$.
Following~\cite{collins2005online}, discriminative features are selected
based on the separation of their class-conditioned feature response distributions $F_{n}^{t}\left(u\right)$ and $B_{n}^{t}\left(u\right)$,
which are 1D histograms extracted for each feature from the object foreground and local background regions, respectively.
Here $u \in {1,...,64}$ is an index into a histogram of feature response values.
In order to extract the object feature response distribution
we use the learned object image from the previous frame $I_{\rm s}^{t-1}$, defined by Eqn.~(\ref{eq:image-generation}),
as a pixel weighting mask:
\begin{equation}
  \label{eq:feature-pmf-generation}
  F_{n}^{t}(u)=\frac{\sum_{\boldsymbol{i}}I_{\rm s}^{t-1}\left(\boldsymbol{i}\right)\delta\left(Z_{n}^{t}\left(\boldsymbol{i}\right)-u\right)}{\Sigma_{u}}\, ,
\end{equation}
where $\delta$ is the Kronecker delta function.
The local background feature response distribution $B_{n}^{t}$ is extracted in a similar way, 
using a weighting mask $1-I_{\rm s}^{t-1}$ over an appropriately sized local image patch.
Using the learned image $I_{\rm s}^{t-1}$ to precisely identify object pixels leads to a more precise extraction of the feature response distributions than with a bounding box (as used in ~\cite{collins2005online}), 
reducing background pollution in the feature learning process~\cite{gatt2010online}.
This, in turn, provides stronger detections for the tracking process.  
This feedback between tracking and feature selection is illustrated in Figure~\ref{fig:system_overview}.

A detection map $\hat{L}_{n}(\boldsymbol{i})$ is computed for each feature by back-projecting its Likelihood Ratio $L_{n}(u)= F_{n}^{t}\left(u\right)/B_{n}^{t}\left(u\right)$
into its feature map and normalizing: $\hat{L}_{n}(\boldsymbol{i})=L_{n}\left(u=Z_{n}^{t}\left(\boldsymbol{i}\right)\right) / \,{\rm max}({L_{n}\left(u=Z_{n}^{t}\left(\boldsymbol{i}\right)\right))}$, see~\cite{collins2005online} for the original formulation and~\cite{gatt2010online} for an illustrated example.
Online feature selection then involves choosing the most discriminable set of $N$ detection maps, 
with $N = 6$ chosen empirically, as similar values ($4-8$) yielded comparable tracking performance.
By considering the feature response in each pixel of a local image region (i.e. object, local background, or both)
as a discrete random variable $\boldsymbol{z}^{t}_{n}$,
we use Maximum Marginal Diversity (MMD)~\cite{vasconcelos2003mmd} to approximate the \emph{infomax space}:
the subset of $N$ features that maximizes its own \emph{mutual information} with the class label random variable $\boldsymbol{c}$.
When applied to feature selection in discriminant tracking~\cite{mahadevan2013saliency,stamatescu2015features}
MMD involves scoring each feature by its mutual information ${\mathcal I}(\boldsymbol{z}_{n}^{t};\boldsymbol{c})$
with the object ($\boldsymbol{c}=1$) and local background ($\boldsymbol{c}=0$) class labels:
\begin{equation}
	{\mathcal I}(\boldsymbol{z}^{t}_{n};\boldsymbol{c}) =  \sum_{c=0}^{1}p(\boldsymbol{c}=c)\mathcal{R}[p(\boldsymbol{z}_{n}^{t}=u|\boldsymbol{c}=c)||p(\boldsymbol{z}_{n}^{t}=u)]\;, \label{eq:mmd}
\end{equation}
where {$\mathcal{R}[p(u)||q(u)]=\sum_{u\in U}p(u)log_{2}\frac{p(u)}{q(u)}$} is the Kullback-Leibler divergence between two distributions $p$ and $q$.
Here the class-conditioned feature response distributions $p(\boldsymbol{z}_{n}^{t}=u|\boldsymbol{c}=1)$ and $p(\boldsymbol{z}_{n}^{t}=u|\boldsymbol{c}=0)$
are given by $F_{n}^{t}(u)$ and $B_{n}^{t}(u)$, respectively, while $p(\boldsymbol{z}_{n}^{t}=u)$ corresponds to the combined object and local background regions.

The most discriminative detection maps are selected by choosing the $N$ highest scores given by Eqn.~(\ref{eq:mmd}),
and these are summed pixel-wise in a weighted average
to produce a fused detection map $I_{\rm m}^{t}$ that serves as input to the tracking algorithm:
\begin{equation}
I_{\rm m}^{t}\left(\boldsymbol{i}\right)=\sum_{n=1}^{N}w_{n}\hat{L}_{n}\left(\boldsymbol{i}\right)\textrm{ .}\label{eq:Likelihood Ratio Fusion}
\end{equation}
The weights in Eqn.~(\ref{eq:Likelihood Ratio Fusion}) are given by $w_{n} =  {\rm I}(\boldsymbol{z}^{t}_{n};\boldsymbol{c}) \times B$,
where the \emph{similarity score} $B$ is the Bhattacharyya coefficient~\cite{bhattacharyya1943distance}:
\begin{equation}
   B= \sum_{u} \sqrt{F_{n, {\rm m}}^{t}\left(u\right) F_{n,{\rm s}}^{t-1}\left(u\right)} \textrm{ ,}\label{eq:sim}
\end{equation}
which rewards temporal consistency between the object feature response $F_{n,{\rm m}}^{t}$
measured in the current frame according to Eqn.~(\ref{eq:feature-pmf-generation})
and an object feature response learned up to the previous frame $F_{n,{\rm s}}^{t-1}$.
The learned posterior feature response $F_{n,{\rm s}}^{t}$ is updated at each frame by
\begin{equation}
   F_{n,{\rm s}}^{t}\left(u\right)=\frac{F_{n,{\rm s}}^{t-1}\left(u\right)F_{n,{\rm m}}^{t}\left(u\right)}{\Sigma_{u}}\textrm{ .}\label{eq:update}
\end{equation}

\subsection{Shape Estimating Filters}

We next address the problem of adaptively learning an object state model, which includes a probabilistic representation of its shape.
The proposed solution is a single object tracker called the Shape Estimating Filter (SEF)~\cite{wong2009shape},
which combines spatiotemporal information from past frames with new measurements
to recursively estimate the object position, velocity and shape.
A SEF autonomously correlates recurring saliency from each new fused detection map into shape and trajectory estimates.

Assuming that only a single object is present in an image, the 2D PMF $I(\boldsymbol{i})$
is used to describe the probability that a given pixel $\boldsymbol{i}=(i_{1},i_{2})$ belongs to that object.
The PMF $I(\boldsymbol{i})$ can then be factored into 2D PMFs for shape $S(\boldsymbol{j})$ and position $X(\boldsymbol{x})$.
Here $X(\boldsymbol{x})$ represents the probability that the object center of mass has position $\boldsymbol{x}=(x_{1},x_{2})$,
while $S(\boldsymbol{j})$ is proportional to the probability that the pixel $\boldsymbol{j}=(j_{1},j_{2})$ is part of the object.
The vectors $\boldsymbol{i}$, $\boldsymbol{j}$ and $\boldsymbol{x}$ are considered 2D random variables operating on the set of integers.
The relationship between image, position and shape random variables is given by $\boldsymbol{i}=\boldsymbol{x}+\boldsymbol{j}$,
which can be expressed as $\boldsymbol{x}=\boldsymbol{i}-\boldsymbol{j}$, or as $\boldsymbol{j}=\boldsymbol{i}-\boldsymbol{x}$.
This relationship allows the shape of an object to be decoupled from its position in the image.

In order to describe the object motion across a sequence of images in an adaptive manner,
2D random variables are used to model acceleration $\boldsymbol{a}$ and velocity $\boldsymbol{v}$.
These variables are described by the PMFs $A(\boldsymbol{a})$ and $V(\boldsymbol{v})$, respectively.
Assuming a simplified Euler motion (non-rotational point-mass) for the object and that $\Delta t=t-(t-1)=1$
leads to the following relationships:
$\boldsymbol{v}^{t}=\boldsymbol{v}^{t-1}+\boldsymbol{a}^{t}$ and $\boldsymbol{x}^{t}=\boldsymbol{x}^{t-1}+\boldsymbol{v}^{t}$.
Rearranging, this gives: $\boldsymbol{v}^{t}=\boldsymbol{x}^{t}-\boldsymbol{x}^{t-1}$ and $\boldsymbol{a}^{t}=\boldsymbol{v}^{t}-\boldsymbol{v}^{t-1}$.

To handle deformable objects, the 2D PMF $R(\boldsymbol{r})$ is defined as the change in shape from one frame to the next,
which is described by the random variable relationship $\boldsymbol{r}=\boldsymbol{j}^{t}-\boldsymbol{j}^{t-1}$.

\begin{figure}[!t]
\centering
\includegraphics[trim = 0mm 30mm 0mm 30mm, clip, width=7.5cm]{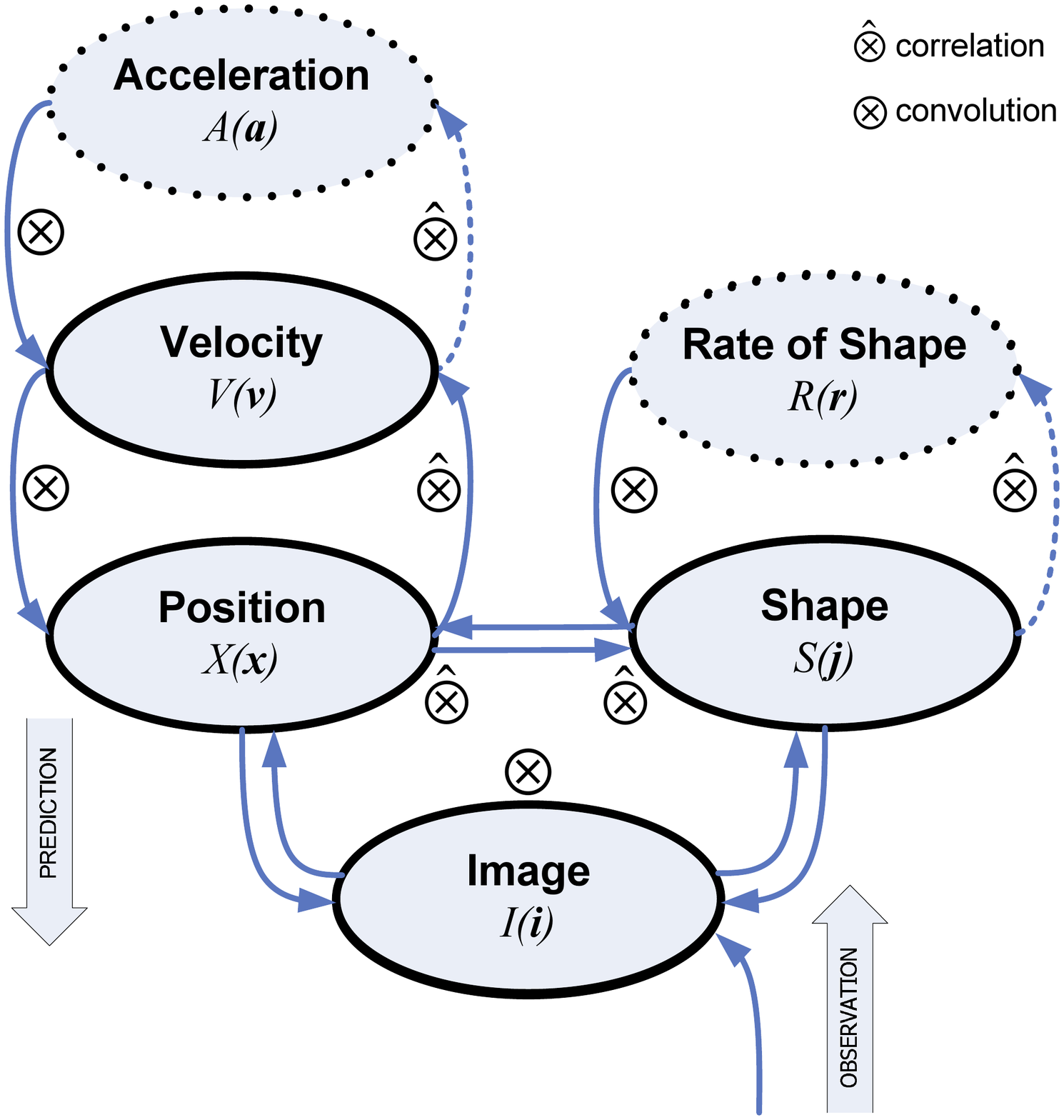}
\caption{\label{sys_sef} The hierarchical state model of the Shape Estimating Filter (SEF)~\cite{wong2009shape}.
  Predictions are propagated from the top-down and new observations from the bottom-up.
  Predictions and observations are combined at each layer to provide an approximate Bayesian update of the state model.
}
\end{figure}

These random variable relationships are used to build the SEF algorithm,
using the operations of convolution $\otimes$ and cross-correlation $\hat{\otimes}$, as illustrated in Figure \ref{sys_sef}.
The SEF state-space hierarchy provides a framework for combining top-down predictions with bottom-up sensory measurements
through a Bayesian update process. 

Predictions are made by traversing down the state model hierarchy (starting in the top left of Figure \ref{sys_sef}) according to:
\begin{align}
\boldsymbol{v}^{t} & =\boldsymbol{v}^{t-1}+\boldsymbol{a}\quad & \Rightarrow\quad V_{\rm p}^{t} & =V{}_{\rm s}^{t-1}\otimes A_{0}\textrm{ ,}\label{eq:predict velocity}\\
\boldsymbol{x}^{t} & =\boldsymbol{x}^{t-1}+\boldsymbol{v}^{t}\quad & \Rightarrow\quad X_{\rm p}^{t} & =X{}_{\rm s}^{t-1}\otimes V{}_{\rm p}^{t}\textrm{ ,}\label{eq:predict position}\\
\boldsymbol{j}^{t} & =\boldsymbol{j}^{t-1}+\boldsymbol{r}\quad & \Rightarrow\quad S{}_{\rm p}^{t} & =S_{\rm s}^{t-1}\otimes R_{0}\textrm{ ,}\label{eq:predict shape}\\
\boldsymbol{i}^{t} & =\boldsymbol{j}^{t}+\boldsymbol{x}^{t}\quad & \Rightarrow\quad I{}_{\rm p}^{t} & =S{}_{\rm p}^{t}\otimes X{}_{\rm p}^{t}\textrm{ ,}\label{eq:predict image}
\end{align}
where $A_{0}$ and $R_{0}$ are 2D Gaussian priors.

Given $I_{\rm m}^{t}$, measurements are made by traversing up the state model hierarchy (starting at the bottom of Figure \ref{sys_sef}) according to:
\begin{align}
\boldsymbol{x}^{t} & =\boldsymbol{i}^{t}-\boldsymbol{j}^{t}\quad & \Rightarrow\quad X_{\rm m}^{t} & =I{}_{\rm m}^{t}\hat{\otimes}S{}_{\rm p}^{t}\textrm{ ,}\label{eq:measure position}\\
\boldsymbol{v}^{t} & =\boldsymbol{x}^{t}-\boldsymbol{x}^{t-1}\quad & \Rightarrow\quad V_{\rm m}^{t} & =X{}_{\rm m}^{t}\hat{\otimes}X{}_{\rm s}^{t-1}\textrm{ ,}\label{eq:measure velocity}\\
\boldsymbol{j}^{t} & =\boldsymbol{i}^{t}-\boldsymbol{x}^{t}\quad & \Rightarrow\quad S_{\rm m}^{t} & =I{}_{\rm m}^{t}\hat{\otimes}X{}_{\rm s}^{t}\textrm{ .}\label{eq:measure shape}
\end{align}

An approximate Bayesian update scheme is used to combine top-down predictions with bottom-up observations.
The posterior PMFs of position, velocity and shape are described by:
\begin{align}
X_{\rm s}^{t}\left(\boldsymbol{x}\right) & =\frac{X_{\rm m}^{t}\left(\boldsymbol{x}\right)X_{\rm p}^{t}\left(\boldsymbol{x}\right)}{\Sigma_{\boldsymbol{x}}}\textrm{ ,}\label{eq:Smoothing-Position}\\
V_{\rm s}^{t}\left(\boldsymbol{v}\right) & =\frac{V_{\rm m}^{t}\left(\boldsymbol{v}\right)V_{\rm p}^{t}\left(\boldsymbol{v}\right)}{\Sigma_{\boldsymbol{v}}}\textrm{ ,}\label{eq:Smoothing-Velocity}\\
S_{\rm s}^{t}\left(\boldsymbol{j}\right) & =\frac{S_{\rm m}^{t}\left(\boldsymbol{j}\right)S_{\rm p}^{t}\left(\boldsymbol{j}\right)}{\Sigma_{\boldsymbol{j}}}\textrm{ .}\label{eq:Smoothing-Shape}
\end{align}

\subsection{Competitive Attention Correlation Tracking using Shape}

Finally, we address the problem of automatically associating new measurements to multiple system tracks.
The proposed solution, which extends the work of Strens and Gregory~\cite{strens2003cat}, operates multiple SEFs simultaneously
in a competitive attentional framework designed to enforce the tracking of multiple objects.
Under this scheme, the SEFs track everything in the scene, including parts of the background or sources of clutter,
so that every new measurement is assigned to the SEF that best describes that measurement~\cite{wong2014modeldrift}.

\begin{figure}
  \begin{center}
    \includegraphics[trim = 12mm 5mm 12mm 5mm, clip, height=4.3cm]{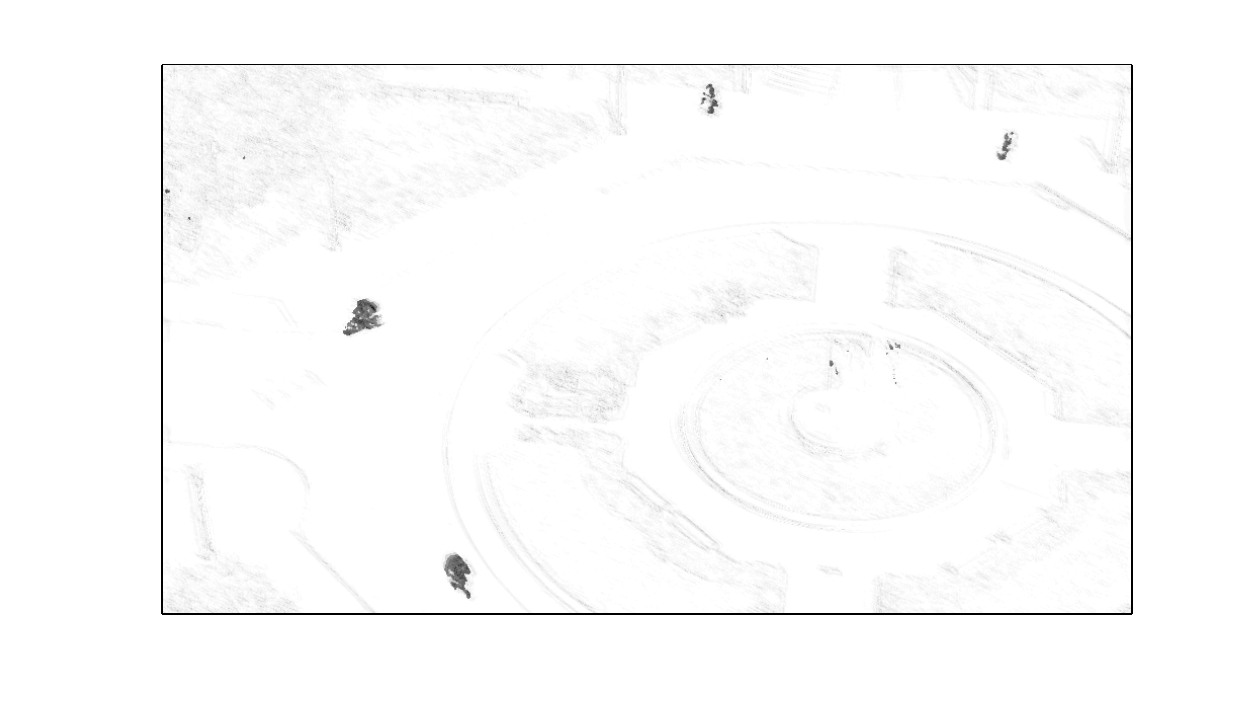}
    \includegraphics[trim = 12mm 5mm 12mm 5mm, clip, height=4.3cm]{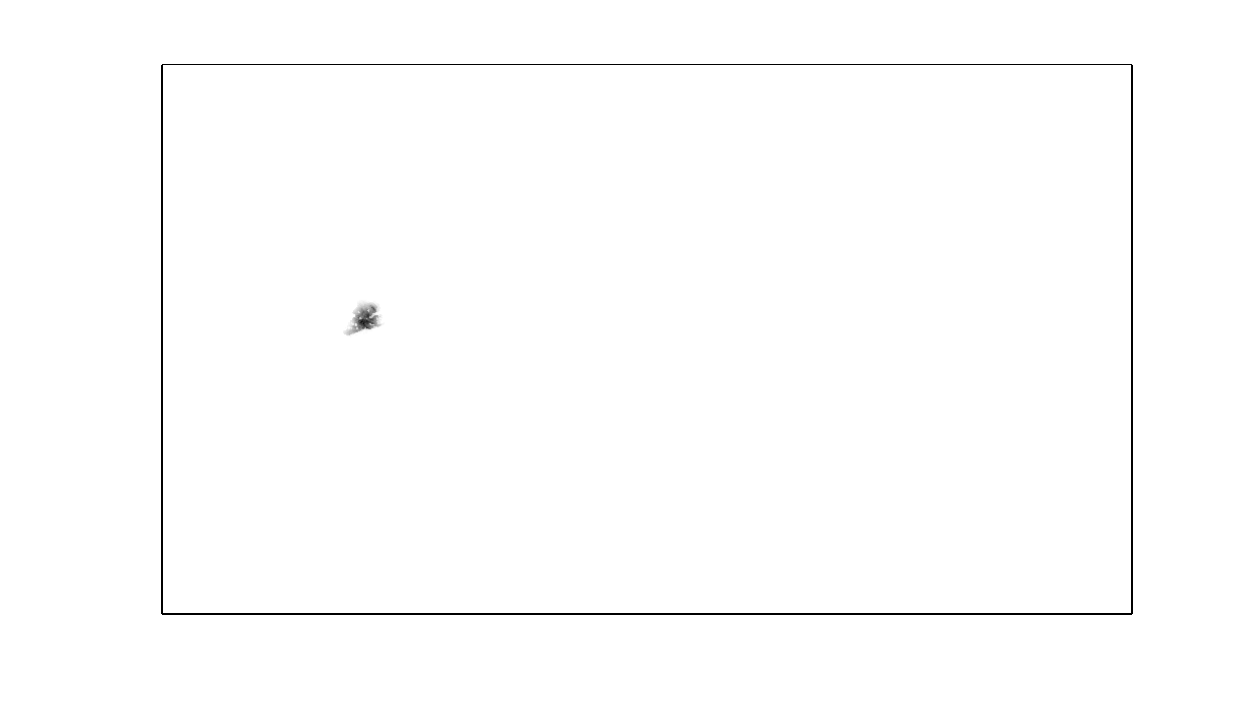}
\end{center}
\caption{ \label{fig:frame61_attention}
  The selective attentional mechanism of {CACTuS-FL}, for SEF $k=33$, which is tracking a cyclist,
  in frame $61$ of Neovision2 Tower image sequence $001$.
  The $I_{\rm m}^{k}\left(\boldsymbol{i}\right)$ fused detection map from Eqn.~(\ref{eq:Likelihood Ratio Fusion}) (top)
  is modulated by the spatial area of attention $\beta^{k}(\boldsymbol{i})$ 
  to form the bottom-up input for the SEF (bottom).
}
\end{figure}

For each frame $t$, the multi-object tracking algorithm operates $k=1,..,K$ individual SEFs.
The bottom-up input of each SEF $k$ is modulated by an association term $\beta^{k}(\boldsymbol{i})$,
so that Eqn.~(\ref{eq:Likelihood Ratio Fusion}) becomes
\begin{equation}
I_{\rm m}^{k}\left(\boldsymbol{i}\right)=\beta^{k}(\boldsymbol{i})\sum_{n=1}^{N}w_{n}\hat{L}_{n}\left(\boldsymbol{i}\right)\textrm{ .}
\end{equation}
As shown in Figure~\ref{fig:frame61_attention}, top-down modulation provides each SEF with a spatial area of attention to collect new measurements.
The term $\beta^{k}(\boldsymbol{i})$ is computed from learned predictions about the expected image: 
\begin{equation}
\beta^{k}\left(\boldsymbol{i}\right)=\frac{I{}_{\rm p}^{k}\left(\boldsymbol{i}\right)}{\sum_{j=1}^{K}I{}_{\rm p}^{j}\left(\boldsymbol{i}\right)}\textrm{ .}
\end{equation}
This selective attentional mechanism modifies the bottom-up input to each SEF,
enabling individual SEFs to selectively ignore pixels that are strongly claimed by another SEF,
where $0 \leq \beta^{k}\left(\boldsymbol{i}\right) \leq 1$ describes the strength of the claim of pixel at location $\boldsymbol{i}$ by SEF $k$.

By assuming a 2D Gaussian prior shape $S_{0}$, an additional attentional mechanism is introduced by replacing Eqn.~(\ref{eq:measure position}) with
\begin{equation}
X{}_{\rm m}^{k}\left(\boldsymbol{x}\right)=I_{\rm m}^{k}\left(\boldsymbol{i}\right)\hat{\otimes}(S_{\rm p}^{k}\left(\boldsymbol{j}\right)S_{0}\left(\boldsymbol{j}\right))\textrm{ .}\label{eq:measure position with attention}
\end{equation}
This introduces a self-centering capability to the system~\cite{cho2005center},
which reduces the problem of \emph{model drift}~\cite{matthews2004modeldrift} that affects correlation trackers~\cite{wong2005act}. 

\begin{figure}
  \begin{center}
    \includegraphics[trim = 12mm 5mm 12mm 5mm, clip, height=4.3cm]{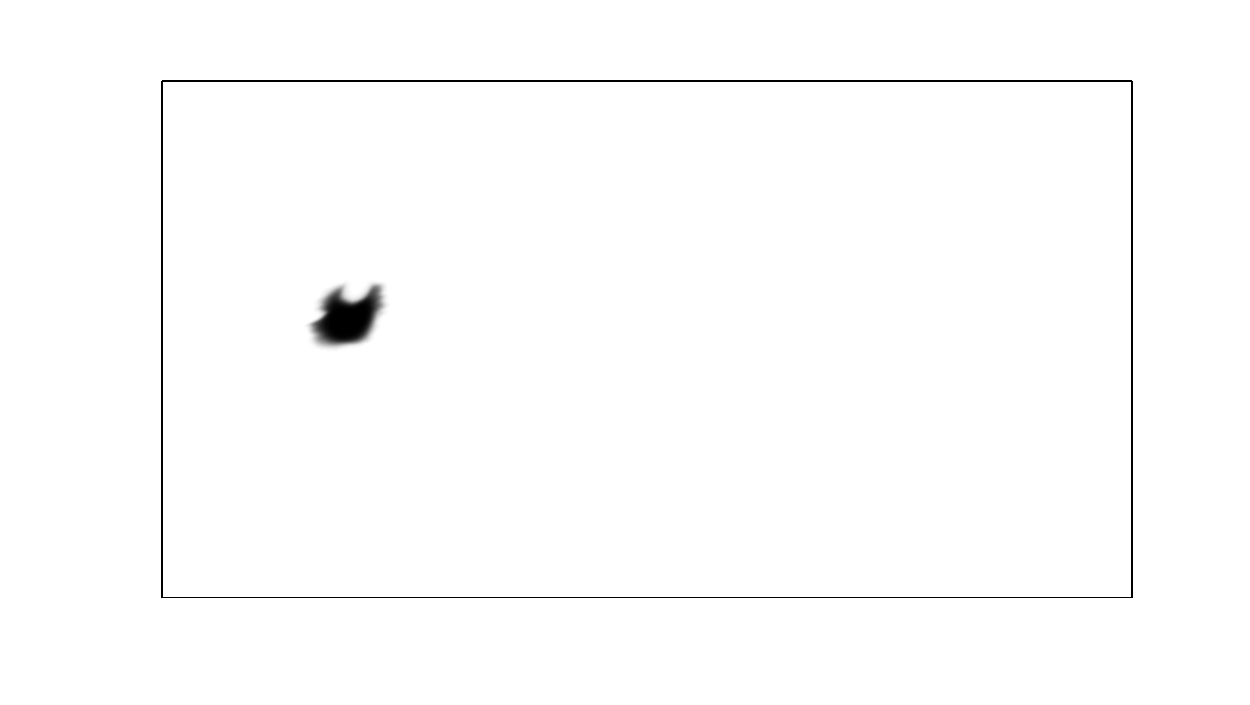}
    \includegraphics[trim = 12mm 5mm 12mm 5mm, clip, height=4.3cm]{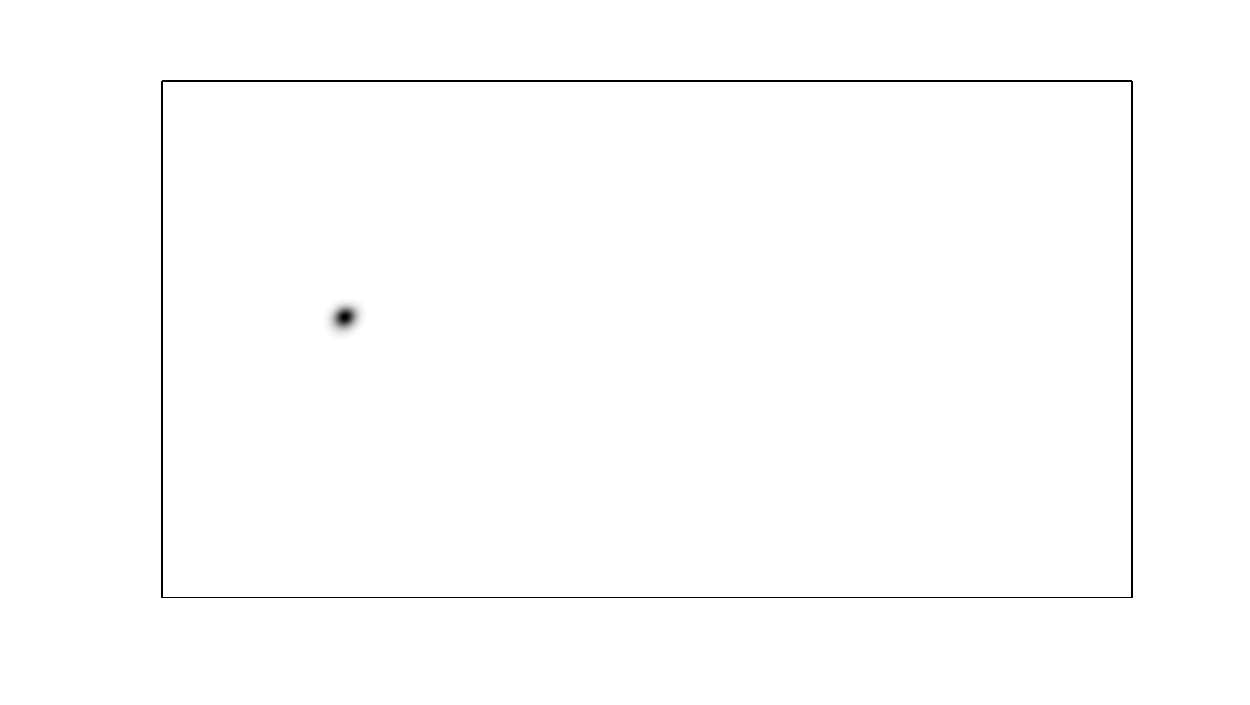}
\end{center}
\caption{ \label{fig:frame61_competition}
  The association probability $C^{k}(\boldsymbol{x})$ (top)
  and the posterior position PMF $X_{\rm s}^{k}\left(\boldsymbol{x}\right)$ (bottom)
  for SEF $k=33$, which is tracking a cyclist,
  in frame $61$ of Neovision2 Tower image sequence $001$.
  }
\end{figure}

In order to encourage SEFs to track multiple objects, Eqn.~(\ref{eq:Smoothing-Position})
is modified by a \emph{winner-take-more} competitive mechanism~\cite{strens2003cat}.
Under this scheme, which has the inherent assumption that different objects tend to occupy different positions,
$K$ separate SEFs compete over position $\boldsymbol{x}$ to track every object in the scene.
Each SEF $k$ competes against all SEFs for its own share of the total association probability 
$\sum_{l=1}^{K}C^{l}(\boldsymbol{x}) = 1$ at each position $\boldsymbol{x}$. 
The individual association probability $C^{k}$, which is shown for a single SEF in Figure~\ref{fig:frame61_competition},
is computed using the predicted position $X_{\rm p}^{k}(\boldsymbol{x})$ according to 
\begin{equation}
C^{k}\left(\boldsymbol{x}\right)=\frac{X_{\rm p}^{k}\left(\boldsymbol{x}\right)}{\sum_{l=1}^{K}X{}_{\rm p}^{l}\left(\boldsymbol{x}\right)}.
\end{equation}
The update of position $X_{\rm s}^{k}\left(\boldsymbol{x}\right)$ for each SEF $k$ in Eqn.~(\ref{eq:Smoothing-Position})
is then modified to include this spatial attention modulation for each SEF
\begin{equation}
X{}_{\rm s}^{k}\left(\boldsymbol{x}\right)=\frac{X_{\rm m}^{k}\left(\boldsymbol{x}\right)X_{\rm p}^{k}\left(\boldsymbol{x}\right)C^{k}\left(\boldsymbol{x}\right)}{\Sigma_{\boldsymbol{x}}}.\label{eq:position-competition}
\end{equation}
An example of $X_{\rm s}^{k}(\boldsymbol{x})$ is shown for a single SEF in Figure~\ref{fig:frame61_competition}.
This mechanism enables the SEF that best describes the position state estimate for a particular object
to converge on a region corresponding to that object and exclude other SEFs from that region.
This competition encourages SEFs to track different objects,
rather than all SEFs converging on the most salient object in the scene.

The shape of the object is observed using the relationship $\boldsymbol{j}=\boldsymbol{i}\ - \boldsymbol{x}$.
First, the best estimate of the object location in the current fused detection map $I_{\rm m}^{k}\left(\boldsymbol{i}\right)$
is extracted from the posterior position PMF $X_{\rm s}^{k}\left(\boldsymbol{x}\right)$ according to
$X_{\rm{smax}}^{k}\left(\boldsymbol{x}\right)= \delta\left(\rm{argmax_1}\left(X_{s}^{k}\left(\boldsymbol{x}\right)\right)-\boldsymbol{x}\right)$, where $\rm{argmax_1}$ returns one maximum.
Next, the PMF $X{}_{\rm smax}^{k}\left(\boldsymbol{x}\right)$ is used to extract
the observed shape $S_{\rm m}^{k}\left(\boldsymbol{j}\right)$ from $I_{\rm m}^{k}\left(\boldsymbol{i}\right)$ using: 
\begin{equation}
S{}_{\rm m}^{k}\left(\boldsymbol{j}\right)=I_{\rm m}^{k}\left(\boldsymbol{i}\right)\hat{\otimes}X{}_{\rm smax}^{k}\left(\boldsymbol{x}\right).\label{eq:shape-measurement}
\end{equation}

The process used to update shape has been adapted from~\cite{wong2005act} as a way to mitigate model drift.
First, the degree of match $\rho$ is computed as the $L^2$ normalized cross-correlation at $\boldsymbol{j}=\left(0,0\right)$ of the measured and predicted shapes, $\rho=\hat{S_{\rm m}}\left(\left(0,0\right)\right) \, \hat{\otimes} \, \hat{S_{\rm p}}\left(\left(0,0\right)\right)$, where $0\leq\rho\leq1$ is a scalar.

Next the parameter $\alpha$ is computed as $\alpha(\rho,\lambda)={H}(\rho-\lambda){\rho}^2$, 
where $H(x)$ is the unit step function and the threshold $\lambda$ acts as
the vigilance parameter~\cite{grossberg1976adaptive} to ensure that very
poor observations are not introduced into memory, see~\cite{wong2005act}
for details. 

\begin{figure}
  \begin{center}
    \includegraphics[trim = 35mm 35mm 35mm 35mm, clip, height=5cm]{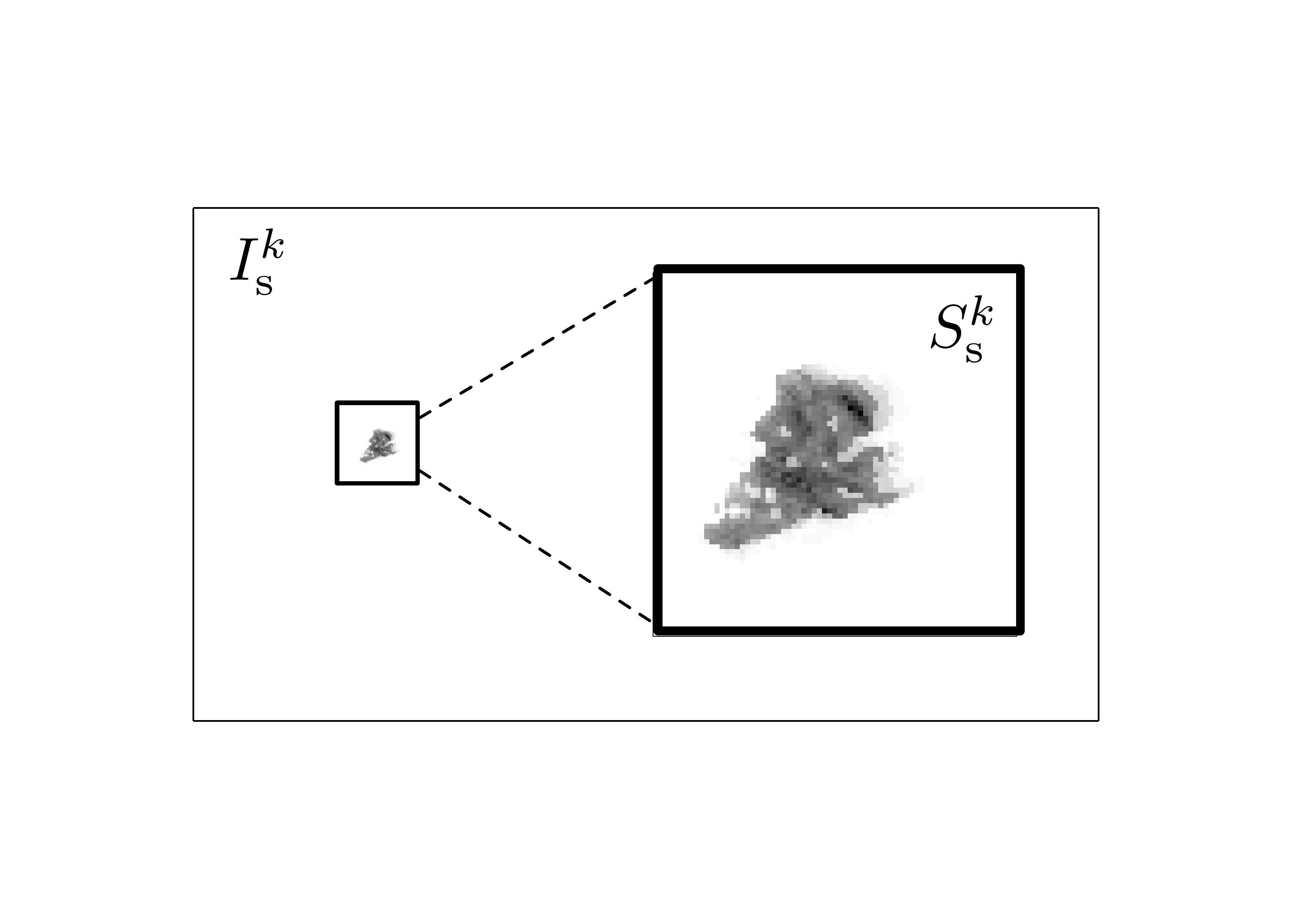}
\end{center}
\caption{ \label{fig:frame61_output_image}
  The posterior image $I_{\rm s}^{k}\left(\boldsymbol{i}\right)$,
  for SEF $k=33$, which is tracking a cyclist,
  in frame $61$ of Neovision2 Tower image sequence $001$.
  The inset shows the corresponding posterior shape $S_{\rm s}^{k}\left(\boldsymbol{j}\right)$.
}
\end{figure}

This controls the degree by which the posterior shape $S_{\rm s}^{k}\left(\boldsymbol{j}\right)$
is influenced by new observations $S_{\rm m}^{k}\left(\boldsymbol{j}\right)$, or prior expectations $S_{\rm p}^{k}\left(\boldsymbol{j}\right)$,
and thus Eqn.~(\ref{eq:Smoothing-Shape}) is replaced with 
\begin{equation}
S_{\rm s}^{k}\left(\boldsymbol{j}\right)=(S_{\rm m}^{k}\left(\boldsymbol{j}\right))^{\alpha}(S{}_{\rm p}^{k}\left(\boldsymbol{j}\right)){}^{(1-\alpha)}\textrm{ .}
\end{equation}
A high degree of match results in a large update of the shape $S_{\rm s}^{k}\left(\boldsymbol{j}\right)$, while a low degree of match leads to a small update.
The resulting posterior shape is shown for a single SEF in Figure~\ref{fig:frame61_output_image}.

Rather than combining the predicted and measured images, the posterior image $I_{\rm s}^{k}\left(\boldsymbol{i}\right)$,
is computed according to a maximum \emph{a posteriori} approach based on the shape and position:
\begin{equation}
I_{\rm s}^{k}\left(\boldsymbol{i}\right)=S_{\rm s}^{k}\left(\boldsymbol{j}\right)\otimes X{}_{\rm smax}^{k}\left(\boldsymbol{x}\right)\textrm{ .}\label{eq:image-generation}
\end{equation}
The posterior image, which is shown for a single SEF in Figure~\ref{fig:frame61_output_image},
then provides top-down guidance for new detections according to Eqn.~(\ref{eq:feature-pmf-generation}) in the object detection stage.

\subsection{Tracking Output}

Each SEF $k$ outputs the posterior image  $I_{\rm s}^{k}$ of the object that it is tracking.
This learned image is then parameterized by calculating its ellipse of second order moments~\cite{hillas195parameters}.
The $2\sigma$ length and width along the ellipse major and minor axis, respectively,
are used to define an oriented output bounding box for each object, in every frame.

\section{What: Object Recognition}
\label{sec:what}

This section describes the S-CNN and SLFN ensemble classification algorithms, 
detailing their supervised offline training and application to online object recognition.

While a variety of image classifiers could act as the \emph{what} processing stream,
S-CNNs and SLFNs, in which only the output layer weights are learned,
have the advantage of being fast to train (on the order of minutes on standard PCs)
and hence are well suited to tasks that require frequent domain-specific re-training.

\subsection{Shallow Convolutional Neural Network}

\subsubsection*{S-CNN Offline Training}

Here we summarise our application-specific S-CNN implementation,
while an in depth description of the algorithm may be found in~\cite{mcdonnell2015overfeat}.
The network architecture consists of five layers:
an input image pixel layer, three hidden unit layers, and an output layer.
Only the weights that project to the final layer are learned.
The S-CNN can be divided into two conceptual stages:
a convolutional filtering and pooling stage formed by the first two hidden layers,
which extract translation and scale invariant features,
and a classification stage consisting of the third hidden layer and the output layer.

{\emph{Stage 1: Convolutional Filtering and Pooling.}}
Each domain-specific S-CNN is trained on a single batch of image patches of size $61 \times 61$ pixels.
The bank of $24$ visual processing filters shown in Figure~\ref{fig:crbm_filters},
which serve as generic object detectors in the \emph{where} processing stream,
are reused here as the first layer of convolutional filters.
Following~\cite{mcdonnell2015overfeat},
the first hidden layer units are obtained by applying a termwise nonlinear function $g_1(u)=u^2$.
The first hidden layer activations are average-pooled and down-sampled
by applying a uniform low pass filter with a pooling size of $18 \times 18$ pixels and stride of $6$ pixels.
Finally a termwise nonlinear function of form $g_2(u)=u^{0.25}$ is applied to obtain the image features.

{\emph{Stage 2: Classification.}}
The features from the second hidden layer are concatenated and linearly projected onto $12000$ hidden units
using a fully-connected set of real-valued input weights, which is set only once during training following the method of~\cite{zhu2015elm}.
Applying the termwise squaring function $g_1(u)$ to every mapped feature yields the third hidden layer activations.
Output labels can then be predicted by linearly mapping these activations using a set of fully-connected output weights
obtained as in~\cite{mcdonnell2015overfeat}, and described further below.

In order to train the S-CNN, pre-processed (see Section III)
$61 \times 61$ pixel image patches are extracted from its training image sequences.
Using the Neovision2 Tower training videos,
this involves extracting image patches from $15$ image sequences ($010-024$)
based on the positions of the ground truth bounding boxes.
To simulate the effect of object tracking during training,
the centre of each patch includes positional Gaussian random jitter about the object ground truth location,
with a standard deviation of $10$ pixels in both the $x$ and $y$ axis directions.
In each patch, the pixels outside a central circular spatial attention region of radius $30$ pixels are set to $0$.
Additional patches are randomly extracted from background regions in each training image
to provide training examples for a background Clutter class.
The training examples are then randomly shuffled and the class abundances are balanced
so that the number of training examples is uniformly spread among four classes: Car, Person, Cyclist and Clutter.

Given that the convolutional filters, the pooling parameters and the classifier input weights are fixed,
the offline training algorithm only involves finding the set of optimal output weights.
These are obtained by forming a set of linear equations
from a single batch of training class labels and output layer activations
and solving for the output weights using least squares regression as in~\cite{mcdonnell2015overfeat}.

\subsubsection*{S-CNN Online Object Recognition}

In online processing, raw pixel image patches, which are centered on the position of each SEF,
are presented as input to the trained S-CNN in the form of an input vector ${\bf x_{\rm test}}$.
Following the matrix notation of~\cite{mcdonnell2015overfeat},
the S-CNN output for each patch is the predicted label vector ${\bf y}_{\rm test}$
whose length corresponds to the number of classes:
\begin{equation}\label{online1}
{\bf y}_{\rm test} = {\bf W}_{\rm out}~g_1({\bf W_{\rm in}}~g_2({\bf W}_{\rm Pool}~g_1({\bf W}_{\rm Filter}{\bf x_{\rm test}}))),
\end{equation}
where the convolution matrices ${\bf W}_{\rm Filter}$ and ${\bf W}_{\rm Pool}$ apply convolutional filtering and pooling, respectively,
the matrix ${\bf W_{\rm in}}$ corresponds to the fully-connected input weights,
and the matrix ${\bf W_{\rm out}}$ corresponds to the fully-connected output weights.
If the S-CNN were used on its own, without applying the SLFN,
the predicted class would be given by the index of the maximum value in ${\bf y}_{\rm test}$.

\subsection{Single Hidden Layer Feedforward Network Ensemble}

\subsubsection*{SLFN Offline Training}

We next train SLFNs to predict the ground truth class label associated with each SEF
by combining object state (\emph{where} stream) information and the corresponding S-CNN (\emph{what} stream) output unit activations.
To reduce the potential for over fitting, an ensemble~\cite{hinton2015distilling} of seven small SLFNs are trained separately. 
Each SLFN employs the same type of architecture as the S-CNN classification stage.
The input features of the first six SLFNs are linearly mapped onto $320$ hidden units
using a fixed set of fully-connected input weights that are set randomly only once in training~\cite{mcdonnell2015elm}.
Using the same approach, the seventh SLFN instead maps the vector form
of the $71 \times 71$ pixel posterior shape (e.g. see Figure~\ref{fig:frame61_output_image}) onto a layer of $12800$ hidden units.
In all SLFN instances, a termwise logistic sigmoid function $g(u) = 1/(1+\exp(-u))$ is applied to each hidden unit,
and these activations are mapped to the output units using an optimal set of fully-connected output weights that is learned during training.
As was done for the S-CNN, the optimal output weights for each SLFN are obtained in one shot using least squares regression.

The training procedure for the first six SLFNs relies on a set of $10$ features,
comprising the \emph{softmax} of the S-CNN output vector from Eqn.~(\ref{online1}) ($6$ features),
and state variables in the form of predicted object bounding box width, length and absolute inclination angle (about the x-axis),
as well as the energy of the posterior position PMF:  $\sum_{\boldsymbol{x}}(X_{\rm s}^{k}\left(\boldsymbol{x}\right))^{2}$,
which measures the degree to which a SEF has collapsed (or latched) onto its object.
Before training the SLFN, each state variable is pre-processed
by subtracting the training sample mean and then normalizing by the \emph{rms} of the entire mean-subtracted training sample,
and these parameters are saved and also used in online pre-processing.
Six SLFNs are then trained using a $65$ dimension input feature vector that is formed by multiplying pairs of features, for all unique pairwise combinations plus the individual unpaired features themselves.

In order to accumulate training examples,
{CACTuS-FL} and the S-CNN are applied the Neovision2 Tower training sequences $001$, $010$, $013$, $014$, and $017$,
for which we added unique object IDs by hand to the original ground truth data.
This allows optimal associations to be made between SEF bounding boxes and ground truth bounding boxes
using the Munkres algorithm~\cite{munkres1957assignment}.
This mapping procedure is used to assign true class labels to each tracked object,
which produces the required set of training labels.
The first six SLFNs are trained by applying a bagging technique that randomly divides the data among six separate sets.
In the case of the posterior shape based (seventh) SLFN, all of the training data is used in a single batch.

\subsubsection*{SLFN Online Object Recognition}

During online processing, given the vector ${\bf f}^{c}_{\rm test}$ of input features appropriate for each trained SLFN $c = 1,...,7$,
the output unit vector ${\bf y'}^{c}_{\rm test}$ is given by:
\begin{equation}\label{online2}
  {\bf y'}^{c}_{\rm test} = {\bf W'}^{c}_{\rm out}~g({\bf W'_{\rm in}}^{c}~{\bf  f_{\rm test}}^{c}),
\end{equation}
where the matrices ${\bf W'_{\rm in}}^{c}$ and ${\bf W'}_{\rm out}^{c}$ correspond to each of SLFN input and trained output weights, respectively.
Finally, a \emph{softmax} function is applied to each output vector,
and the SLFNs are used in an  ensemble by combining their output through an element-wise sum:
\begin{equation}\label{online3}
  {\bf y'}_{\rm ensemble} = \sum_{c=1}^{7} {\rm softmax}({\bf y'}^{c}_{\rm test}).
\end{equation}
The class predicted by the online object recognition system is given by the index of the maximum value in ${\bf y'}_{\rm ensemble}$.

\section{Experimental Evaluation}
\label{sec:evaluation}

This section describes the data used in our experiments
together with a summary of previous evaluations of the main system components.
The section also details our experimental parameters, highlighting any use of prior knowledge,
as well as explaining the performance evaluation metrics.
The system performance is then compared against existing
online object recognition benchmark results~\cite{katsuri2014neuromorphicvision},
while the impact of the main components ({CACTuS-FL}, S-CNN, SLFN ensemble) on its performance is also investigated.

\subsection{Previous Experiments}

We summarise previous experimental results for key components of our online object recognition system:
generic feature extraction, {CACTuS-FL} and the S-CNN, using separate visual tracking and image classification benchmarks.

\subsubsection*{Generic feature extraction}

The choice of convolutional filter bank and individual filter size were made based on experiments~\cite{stamatescu2016filters}
using the Neovision2 Tower training sequence $001$,
where the multi-object tracking performance for all object classes was evaluated
in terms of the best Recall ($ 60.37\%$) and tracking precision MOTP ($ 43.44\%$).

\subsubsection*{Where--CACTuS-FL}

{CACTuS-FL} was evaluated using 8 videos from the VOT2013 single object tracking benchmark~\cite{Kristan_2013_ICCV_Workshops}.
In these experiments~\cite{wong2014modeldrift} the robustness of the tracker was measured by the number of tracking failures.
{CACTuS-FL} incurred 4 tracking failures, as compared to the well known TLD algorithm~\cite{kalal2012tracking}
that had 39 tracking failures, and the state-of-the-art LGT algorithm~\cite{cehovin2013robust} that had 2.75 tracking failures.
A qualitative evaluation on multi-object tracking using soccer videos was also presented.

\subsubsection*{What--S-CNN}

The S-CNN was previously evaluated~\cite{mcdonnell2015overfeat} on the MNIST~\cite{lecun1998mnist}, NORB~\cite{lecun2004norb},
SVHN~\cite{netzer2011svhn} and CIFAR-10~\cite{krizhevsky2009learning} benchmark data sets,
achieving image classification error rates of $0.37\%$, $2.21\%$, $3.96 \%$ and $24.14 \%$, respectively.
In the case of MNIST and NORB, this represents state-of-art image classification accuracy
if excluding techniques that perform training set data augmentation~\cite{wong2016augmentation}.
Furthermore, the experiments showed that S-CNNs are robust
in the sense that the same network metaparameters can be applied across different data sets
to yield similar performance to that obtained by tuning the metaparameters for each data set.

\subsection{Online Object Recognition Experiments}

While the key aspects of our online object recognition system have been tested separately,
testing the integrated system requires a MOT data set with multiple target classes.
As outlined in Section~\ref{sec:review}, existing MOT datasets only exercise tracking of a single class,
and often provide pre-computed detections~\cite{leal2015motchallenge,milan2016mot16}.
By contrast, we require a multi-object, multi-class benchmark
and this is provided by Neovision2~\cite{neovision2}.
This set of challenging image sequences, captured under varying environmental conditions,
contains numerous targets, including stationary objects,
which can undergo occlusions by neighbouring objects or background clutter.

\subsubsection*{Benchmark Data}
The Neovision2 Tower data set consists of $50$ training and $50$ test videos captured from an elevated camera.
In both Tower training and test sets the camera is rotated by $90^{\circ}$ after the first $24$ videos,
and, given that this changes the ground sample distance (pixel/m),
we limit our study to videos $001-024$ in both the training and test sets.

Each image sequence was recorded at $29.97$~${\rm frames}/{\rm s}$ and has $871$ annotated frames,
with ground truth data consisting of a class label and oriented bounding box coordinates for each object of interest.
Five target object classes are present in the Tower data domain (Car, Truck, Bus, Person, Cyclist)
and, through random sampling of the background, we include a sixth Clutter class in order to identify SEFs that are tracking background objects.
Due to the scarcity of Truck and Bus training examples, however, we avoid training and testing on the (few) videos that do contain these object types,
which leaves the following four classes: Car, Person, Cyclist, Clutter.
Following these criteria and also simply excluding any video found to have clearly incorrect ground truth annotations,
we select $12$ Neovision2 Tower test set videos: $001$, $002$, $009$, $010$, $012$, $013$, $017$, $018$, $019$, $021$, $022$, $023$.
This set of videos, which contains 82139 ground truth objects across 10452 image frames, was tested only once.

\subsection{Experimental Parameters}

\begin{table*}[t]
\begin{centering}
  \caption{\label{tab:prior_knowledge} {Prior knowledge.}
  }
\par\end{centering}
\centering{}%
\begin{tabular}{l*{2}{c}r}
  Parameter              & Symbol & Value & Justification \\
\hline
\hline
Size of learned CRBM convolutional filter & & $16 \times 16$ pixels & Tuned for tracking performance of all objects in the scene~\cite{stamatescu2016filters}.\\
\hline
Size of posterior shape & & $71 \times 71$ pixels & Chosen by eye to ensure that $S_{\rm s}^{k}\left(\boldsymbol{j}\right)$ is large enough\\
 & &  & to encompass and collapse on any person, cyclist or car.\\
\hline
Total number of SEFs  & $K$ & 112 & Chosen to encourage competition between SEFs over the entire scene.\\
\hline
Size of second order moment ellipse & & $2\sigma$ & Chosen so that the bounding boxes of collapsed SEFs\\
& &  & encompass their object, but do not extend too far beyond this.\\
\hline
Size of image patch & &  $61 \times 61$ pixels & Chosen so that the S-CNN input captures a large fraction of cars,\\
& &  & but limits the extent of the background around people or cyclists.\\
\hline
\end{tabular}
\end{table*}

\subsubsection*{Prior Knowledge}
While the majority of architectural decisions and run time parameter settings for our system
were chosen empirically based on previous experiments~\cite{wong2005act,wong2014modeldrift,mcdonnell2015overfeat},
some were tuned for the Neovision2 Tower training data set.
These system parameters constitute domain-specific prior knowledge and are listed in Table~\ref{tab:prior_knowledge},
which outlines the reason behind each choice.

\subsubsection*{System Initialization}
{CACTuS-FL} is initialized in the first frame of an image sequence
by positioning the SEFs at regular intervals in a $14 \times 8$ rectangular grid across the scene.
The position, shape and velocity PMFs for each SEF are initialized using isotropic 2D Gaussian distributions.

\subsection{Performance Evaluation Metrics}

The Neovision2 object recognition performance metrics~\cite{ekambaram2012nmotda}
are based on the degree of spatial overlap $d_{t,i,k}$
between each ground truth bounding box region ${\bf r}_{t,i}^{GT}$
and every candidate bounding box region ${\bf r}_{t,k}^{SEF}$ output by the $k^{th}$ SEF:
\begin{equation}
  d_{t,i,k} = \frac{{\bf r}_{t,i}^{GT} \cap {\bf r}_{t,k}^{SEF}}{{\bf r}_{t,i}^{GT} \cup {\bf r}_{t,k}^{SEF}},\label{eq:overlap}
\end{equation}
where $t$ refers to the image frame and $i$ is the ground truth index.

To evaluate the online object recognition performance
we use the publicly available Neovision2 evaluation tool~\cite{ekambaram2012nmotda}.
This uses the Munkres algorithm~\cite{munkres1957assignment} to find optimal SEF to ground truth bounding box associations in each frame
for a spatial overlap threshold of $T_d = 0.2$.
For each image sequence $s$, the system performance in detecting each target object class $o$ (i.e. Car, Person, Cyclist)
is measured using the Normalized Multiple Object Thresholded Detection Accuracy (NMOTDA):
\begin{equation}
\textrm{NMOTDA}_{s,o} =  1 - \frac{\sum_t{(\text{FN}_{t,o} + \text{FP}_{t,o})}} {\sum_t{\text{GT}_{t,o}}},
\label{eq:nmotda}
\end{equation}
where in each frame $t$, $\text{GT}_{t,o}$, $\text{FN}_{t,o}$ and $\text{FN}_{t,o}$
are the number of ground truth objects, false negatives, and false positives, respectively, of object class $o$.
NMOTDA is reported as a number  in the range $(-\infty, 1]$.
The ${\rm NMOTDA}_{s,o}$ scores are then aggregated across all image sequences
to yield the Weighted Normalized Multiple Object Thresholded Detection Accuracy (WNMOTDA):
\begin{equation}
\text{WNMOTDA}_{o} = \frac {\sum_s{\text{NMOTDA}_{s,o}} \times \text{GT}_{s,o}} {\sum_s{\text{GT}_{s,o}}},
\label{eq:wnmotda}
\end{equation}
where the weight $\text{GT}_{s,o}$ is the total number of ground truth objects belonging to class $o$ that are present in image sequence $s$.
Average NMOTDA and Average WNMOTDA are also calculated for all object types
according to Eqn.~(\ref{eq:nmotda}) and Eqn.~(\ref{eq:wnmotda}), respectively, by ignoring the object class label $o$.

\begin{figure}
  \begin{center}
  \includegraphics[trim = 0mm 25mm 0mm 25mm, clip, height=4.5cm]{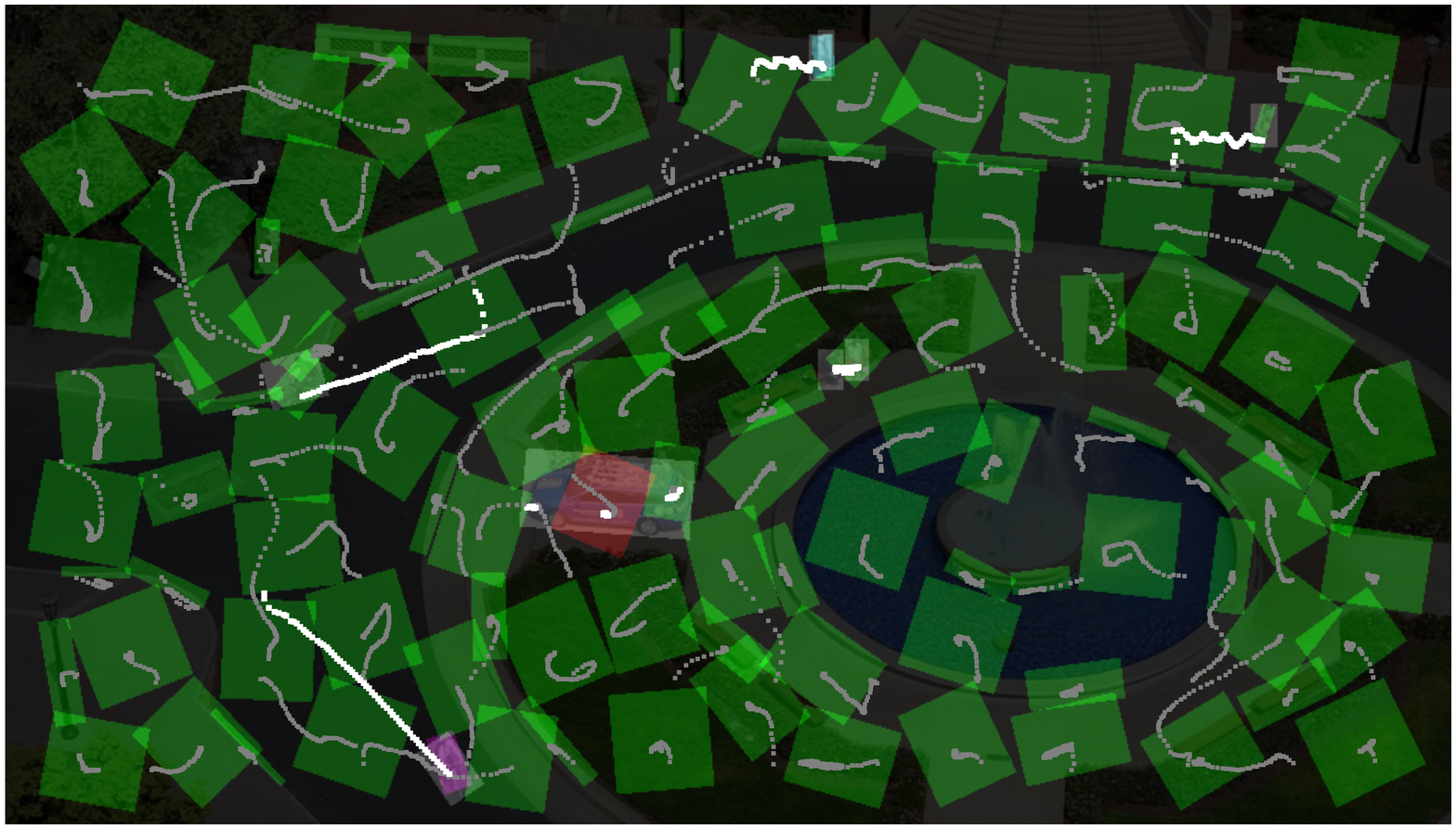}
  \includegraphics[trim = 0mm 25mm 0mm 25mm, clip, height=4.5cm]{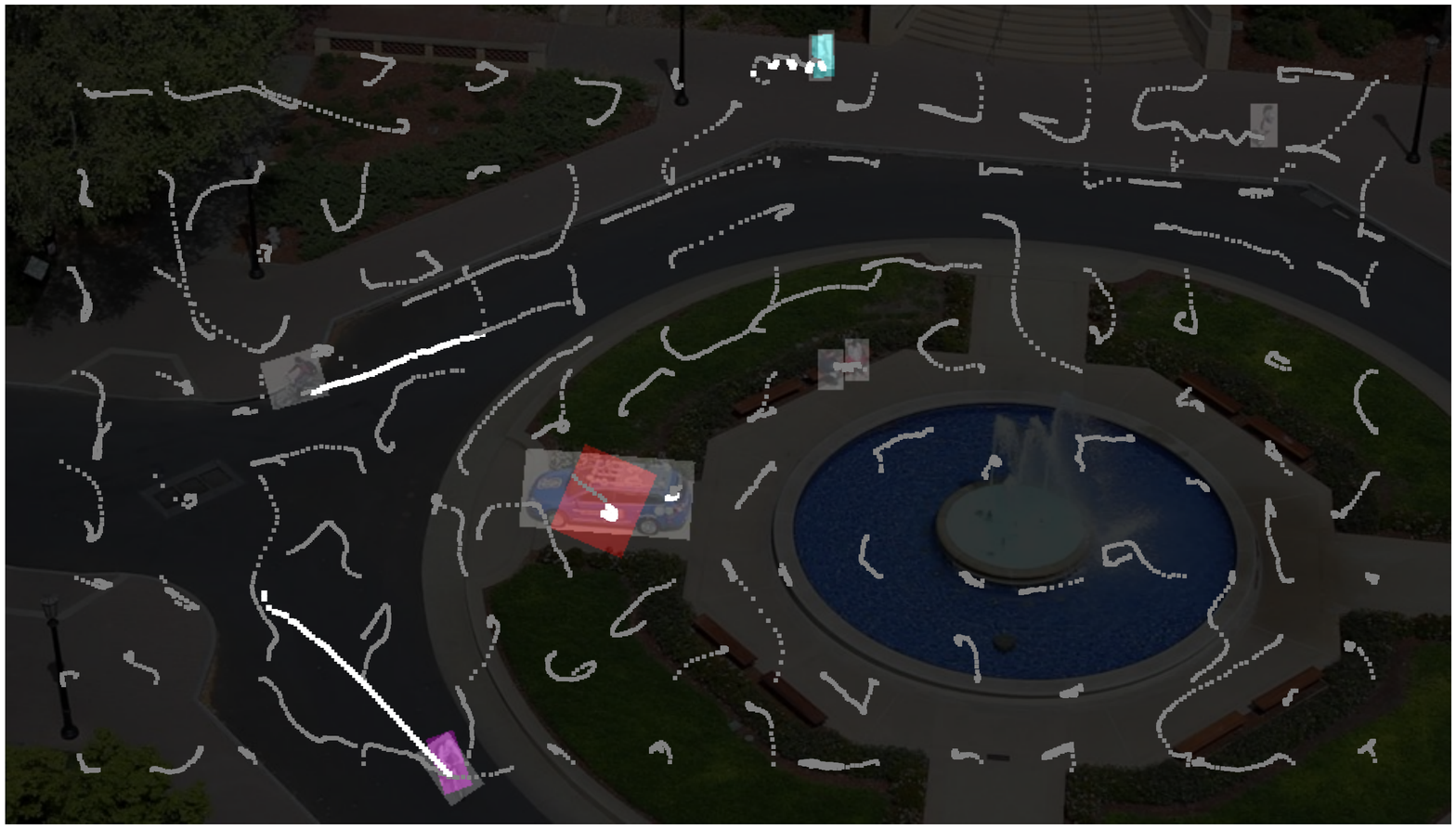}
\end{center}
\caption{ \label{fig:metrics_disp}
  Neovision2 Tower image sequence $001$ frame 61
  showing SEF output from the current frame and CLEAR MOT~\cite{bernardin2008clearmot} tracks up to and including this frame.
  {CACTuS-FL} SEF tracks are shown as grey dots and those identified as SEF and ground truth matches are shown as white dots.
  Ground truth bounding boxes are indicated by shaded grey rectangles, which are centered on every car, person or cyclist in the scene.
  Bounding boxes estimated by {CACTuS-FL} in the current frame, which are computed by parameterizing the object shape learned by each SEF,
  are shown in green, red, magenta and cyan for SEFs classified as Clutter, Car, Cyclist and Person, respectively.
  The top plot shows all bounding boxes, while the bottom plot shows only those bounding boxes that have not been classified as Clutter.}
\end{figure}

In sequences for which we have added ground truth object IDs, such as $001$,
we also apply the CLEAR MOT multi-object tracking metrics~\cite{bernardin2008clearmot}.
Following the implementation of~\cite{andriyenko2012discretecontinuous},
the optimal mapping between SEFs and ground truths is found across all frames in terms of the total spatial overlap.
The associated ground truth and SEF pairs are then identified as \emph{matches} {$j\equiv(i,k)$}
when $d_{t,j}$ exceeds a user-defined threshold $T_d$, which can be varied between $0$ and $1$.
Figure \ref{fig:metrics_disp} illustrates some examples of matched SEF/ground truth pairs for $T_d = 0.2$.
This procedure is used to assign ground truth class labels to SEFs for the purpose of generating SLFN training data.

\subsection{Results}

Table~\ref{tab:results_scnn} lists the training and validation classification accuracies obtained by applying the S-CNN
to image patches extracted around clutter and randomly jittered ground truth object positions.
The un-jittered validation set accuracies obtained here on training video $001$
are comparable to the range of accuracies ($96.77\%-100\%$) obtained by a deep CNN~\cite{cao2015spiking} on Neovision2 Tower data.
The validation results in Table~\ref{tab:results_scnn} indicate that the classification accuracy of the S-CNN degrades considerably,
especially for the Person class, when random position jitter is applied to the image patches,
despite the fact that the same approach was used for the training patches.

\begin{table}[!htbp]
\begin{centering}
  \caption{\label{tab:results_scnn} {S-CNN classification accuracy
    for Tower training ($010-024$) and validation ($001$) sequence image patches.}
  }
\par\end{centering}
\centering{}%
\begin{tabular}{l*{2}{c}r}
  Data set              & Training ($010-024$) & Validation ($001)$ & Validation ($001$) \\
                        & with jitter & without jitter & with jitter \\
\hline
\hline
Car                     & 99.89\% & 100.00\% & 99.89\% \\
\hline
Person                  & 96.76\% & 95.45\% & 78.42\% \\
\hline
Cyclist                 & 95.52\% & 99.46\% & 96.07\% \\
\hline
Clutter                 & 99.07\% & 97.61\% & 97.67\% \\
\hline
\end{tabular}
\end{table}

\begin{figure}
  \begin{center}
  \includegraphics[trim = 0mm 1mm 0mm 1mm, clip, height=6.8cm]{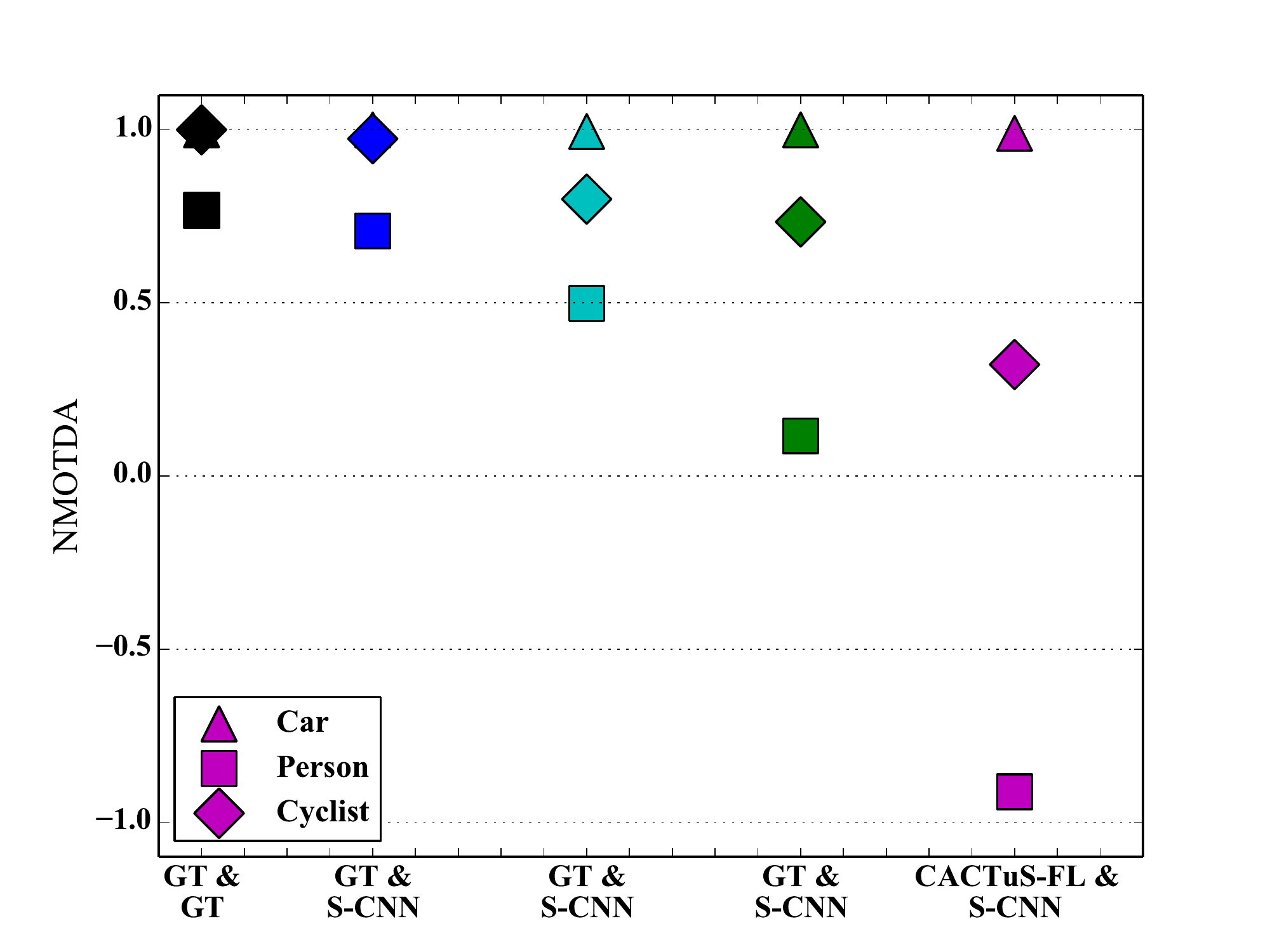}
  \includegraphics[trim = 0mm 1mm 0mm 1mm, clip, height=6.8cm]{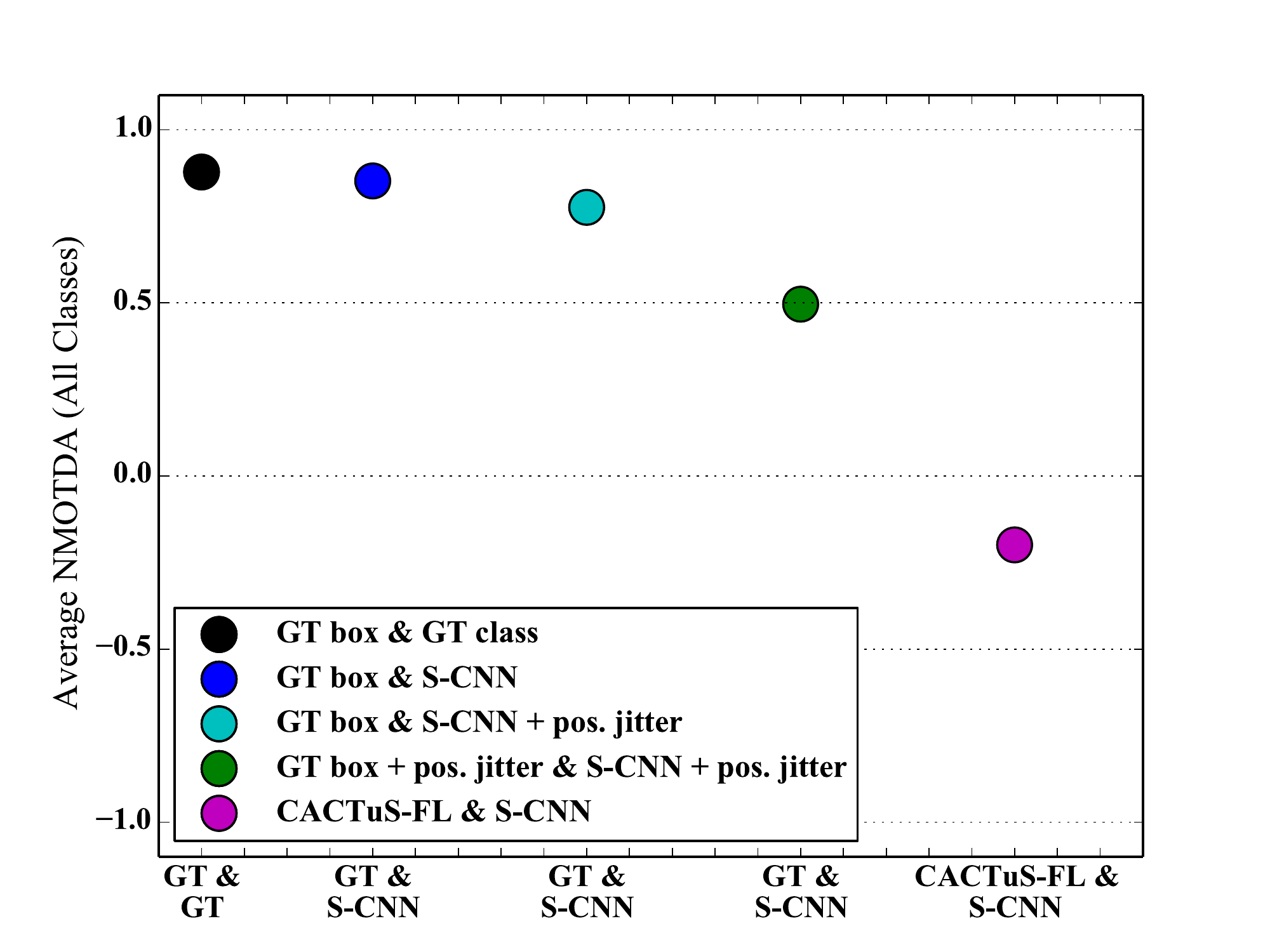}
\end{center}
\caption{ \label{fig:metrics_scnn}
  NMOTDA scores for Neovision2 Tower training sequence $001$, used here for validation.
  The top plot shows NMOTDA for each class,
  the bottom plot shows the Average NMOTDA scores for all classes.
  Black markers correspond to taking the ground truth position and class label data as system outputs (e.g. perfect object tracking and classification).
  Blue markers correspond to perfect tracking and S-CNN based classification.
  Cyan markers correspond to perfect tracking, but here the S-CNN has random jitter applied to its input image patch positions.
  Green markers show the case when position jitter is also applied to the ground truth bounding box to simulate the effect of imperfect tracking.
  In all cases mentioned thus far, the number of SEFs operating in a frame is equal to the number of ground truths in that frame.
  Magenta markers correspond to tracking using {CACTuS-FL} and classification using the S-CNN,
  and in this case $112$ SEFs operate in each frame, with the vast majority tracking background clutter objects.}
\end{figure}

In order to gain some intuition into the impact of tracking and classification accuracy on NMOTDA,
we attempt to decouple the two effects in Figure~\ref{fig:metrics_scnn}, which shows validation results from Tower training sequence $001$.
Starting with perfect tracking and classification, NMOTDA is made progressively worse by first classifying using the S-CNN,
by next adding position jitter in its input images patches, and finally by also adding position jitter to the bounding boxes.
Aside from object tracking accuracy, a second key aspect is that none of these four simulated tests
incorporate clutter-tracking SEFs, which would provide additional false positives.
CACTuS-FL and the S-CNN have the lowest score in Figure ~\ref{fig:metrics_scnn} for this very reason:
operating 112 SEFs across the scene means that the vast majority of SEFs track clutter sources.
The S-CNN on its own, with a typical Clutter class accuracy of $\sim 97.6\%$ (see Table~\ref{tab:results_scnn}),
would then yield $\sim 2.5$ false positives per frame and thus reduce the NMOTDA score.

\begin{figure}
  \begin{center}
  \includegraphics[trim = 0mm 1mm 0mm 1mm, clip, height=6.8cm]{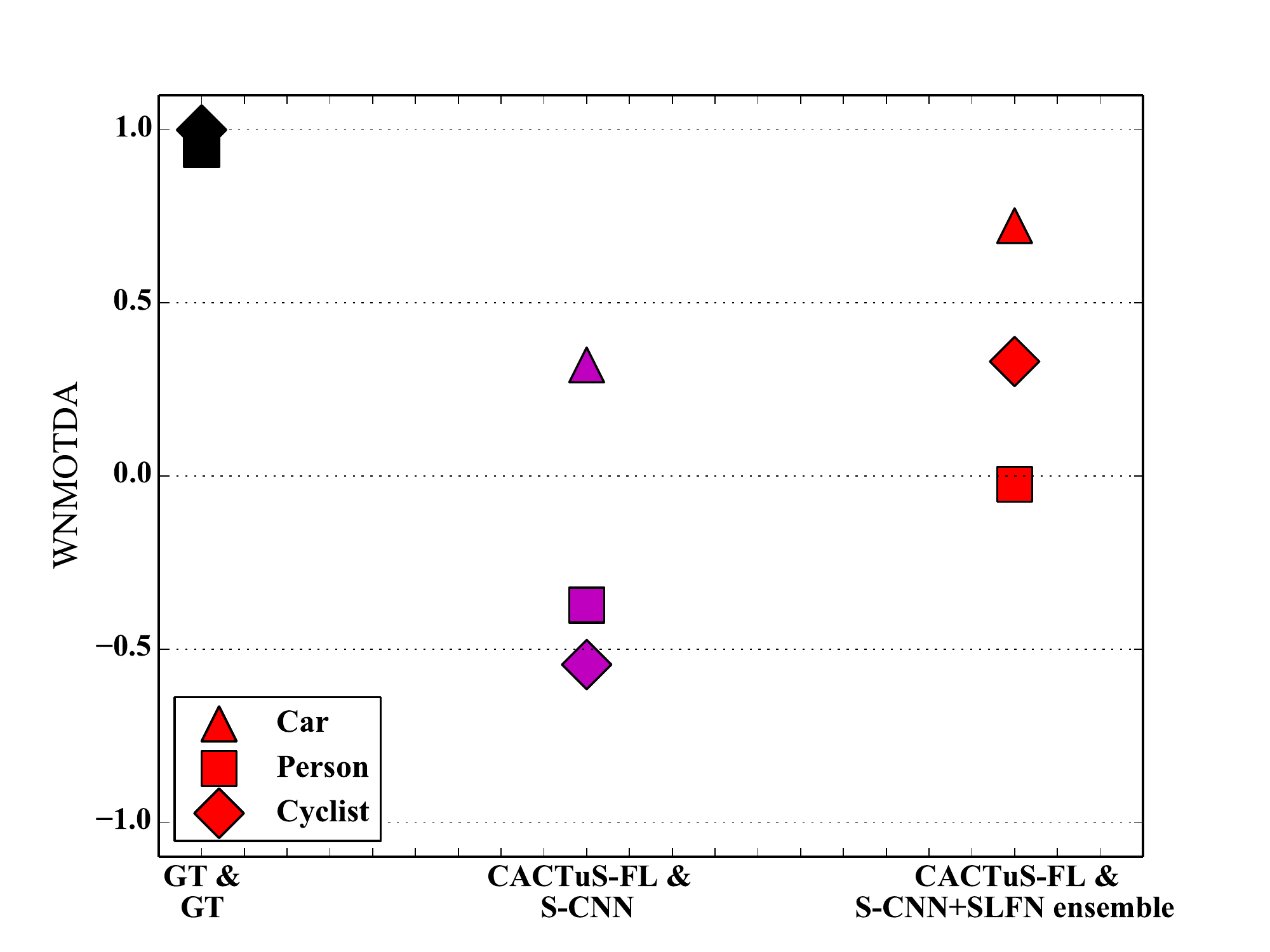}
  \includegraphics[trim = 0mm 1mm 0mm 1mm, clip, height=6.8cm]{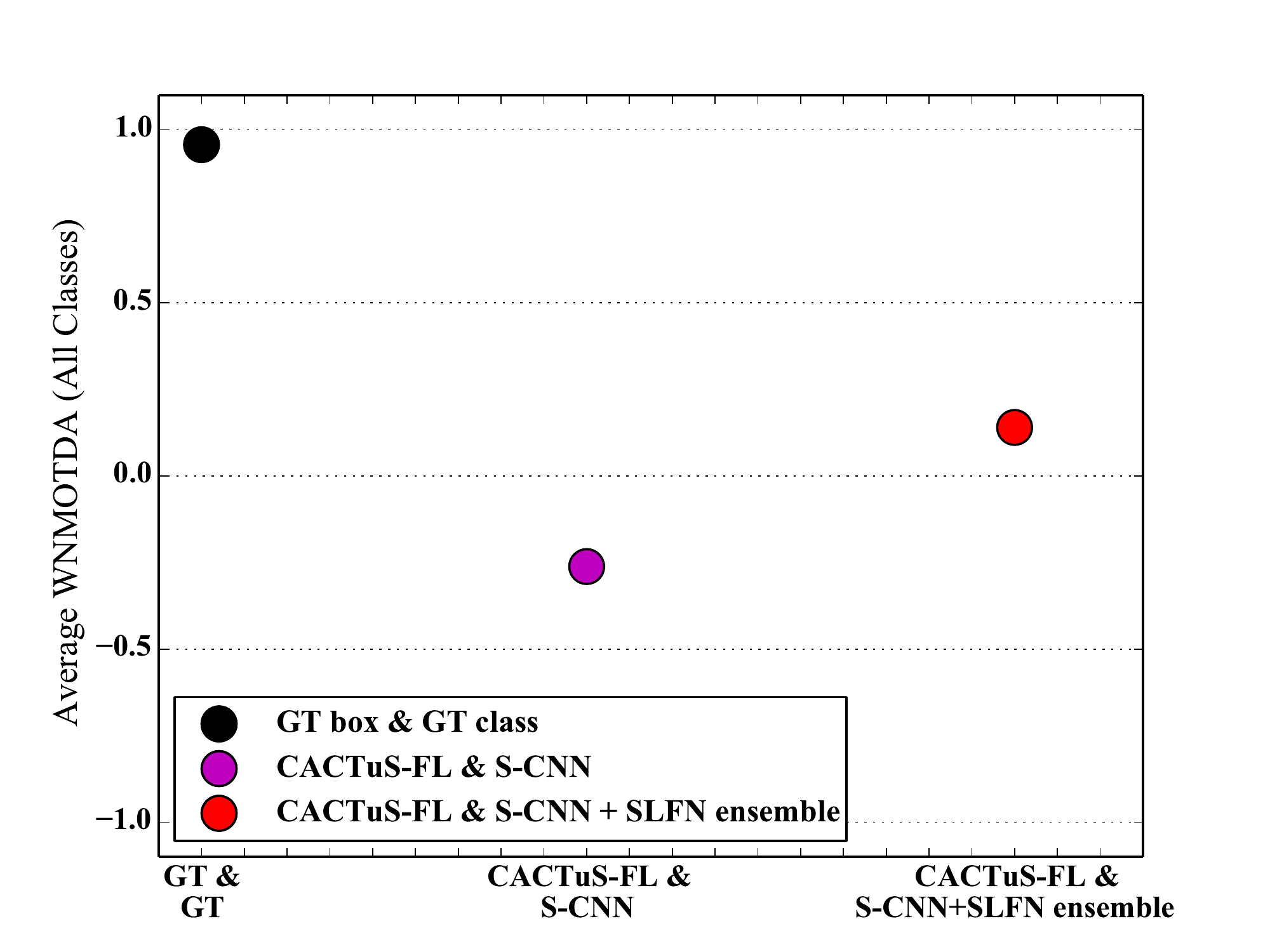}
\end{center}
\caption{ \label{fig:metrics_2nd}
  Neovision2 Tower test WNMOTDA scores, computed across $12$ videos.
  Black markers indicate perfect tracking and classification,
  magenta markers indicate results from {CACTuS-FL} \& S-CNN,
  and red markers indicate results from {CACTuS-FL} \& S-CNN $+$ SLFN ensemble.
  The top plot shows the individual class WNMOTDA,
  while the bottom plot shows the Average WNMOTDA for all classes.}
\end{figure}

This inherent challenge posed by \emph{tracking everything}  motivates the need for a SLFN ensemble.
The Tower test results in Figure~\ref{fig:metrics_2nd} illustrate this point,
where the inclusion of the SLFN ensemble greatly improves both the overall and class-wise performance.
The marked improvement is due to a large reduction in false positives while the number of false negatives tends to remain about the same.
Together, the S-CNN and SLFN ensemble fulfil the dual roles of
($1$) object detection: rejecting Clutter objects while retaining target objects,
and ($2$) object recognition: correctly classifying the target objects (Car, Person, Cyclist),
as illustrated by Figure~\ref{fig:metrics_disp}.

\begin{figure}
  \begin{center}
  \includegraphics[trim = 0mm 1mm 0mm 1mm, clip, height=6.9cm]{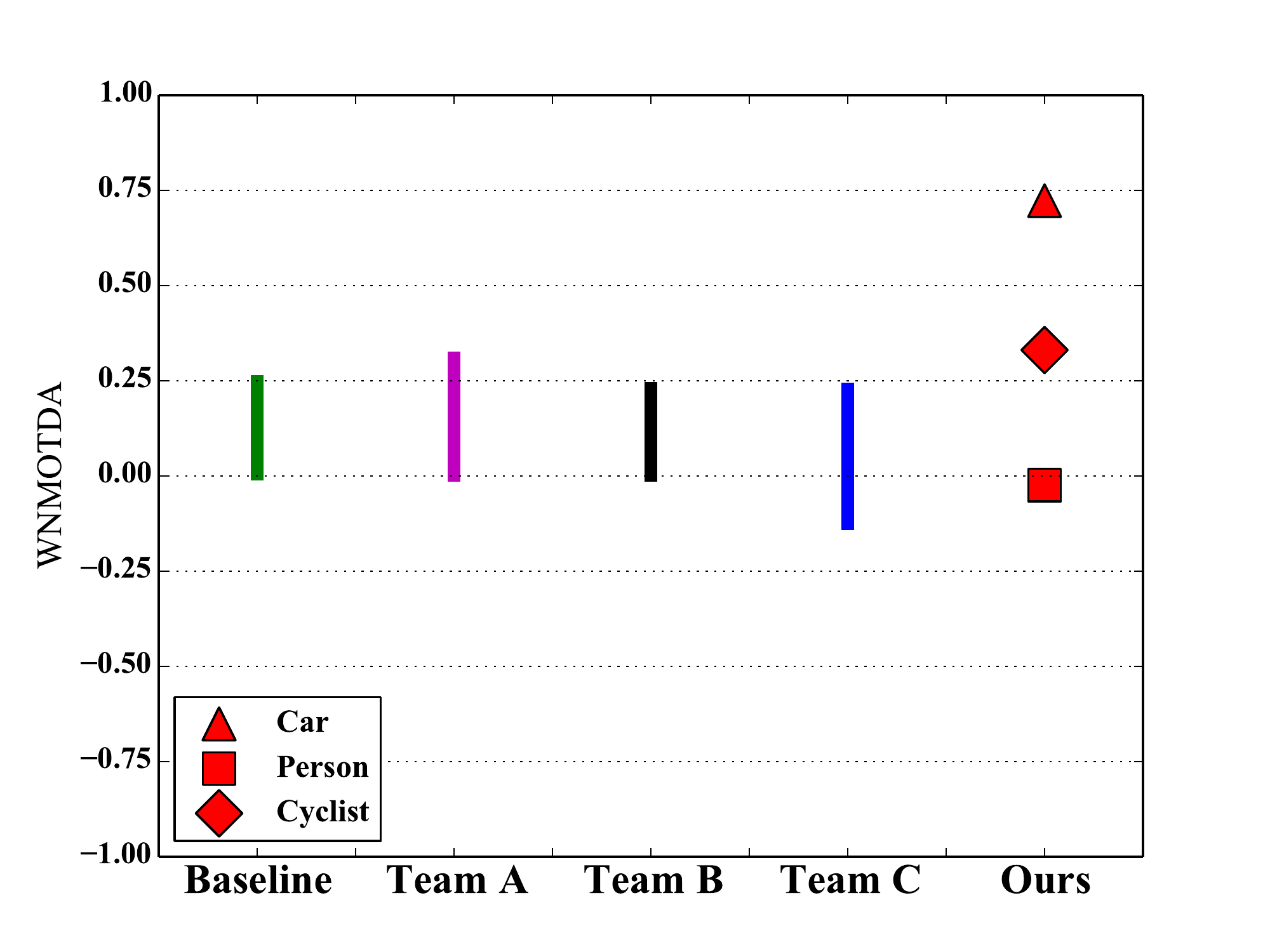}
  \includegraphics[trim = 0mm 1mm 0mm 1mm, clip, height=6.9cm]{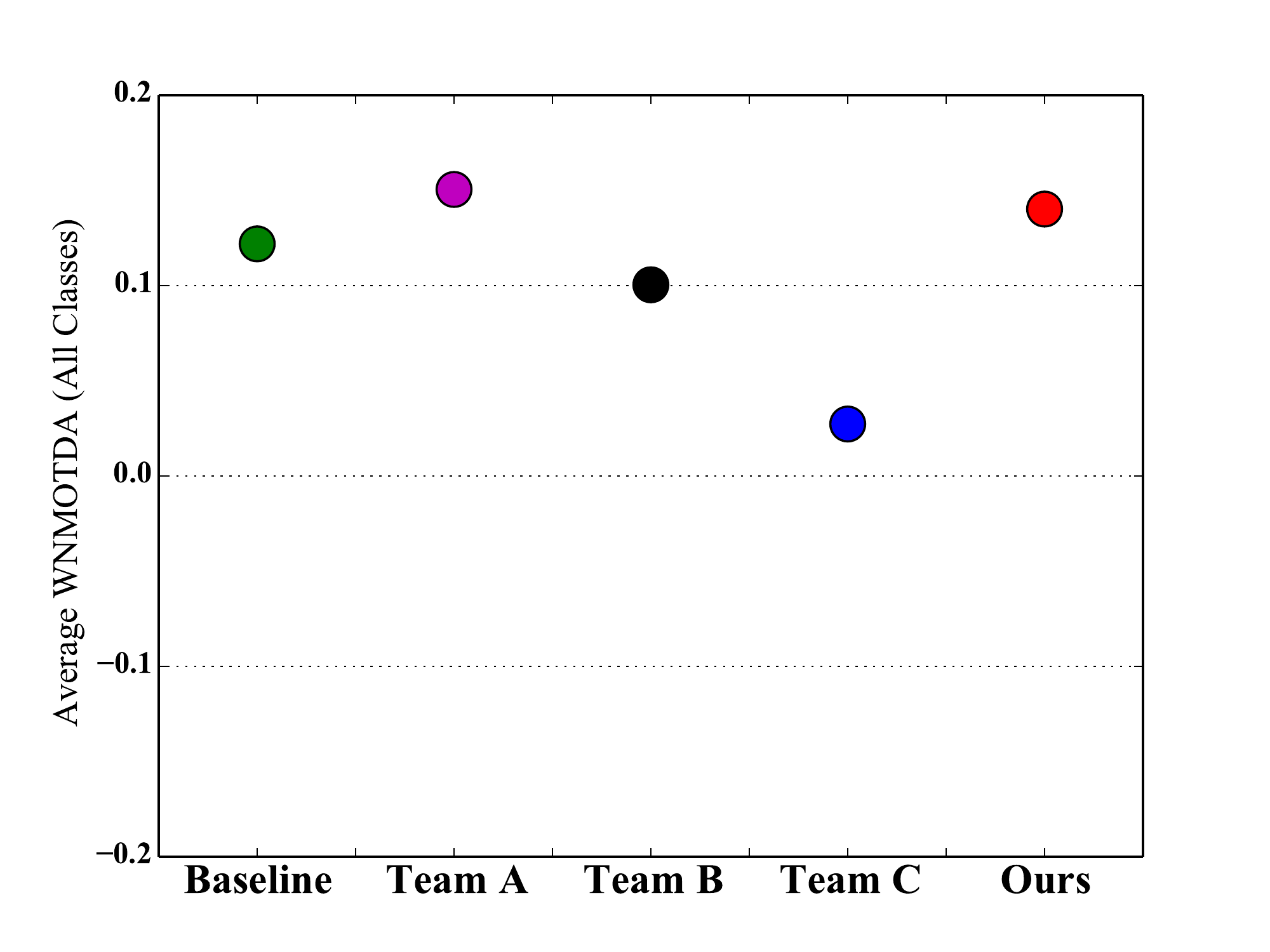}
\end{center}
\caption{ \label{fig:metrics_compare}
  WNMOTDA published by other teams~\cite{katsuri2014neuromorphicvision} (Baseline and Teams A$-$C)
  and WNMOTDA obtained with our system ({CACTuS-FL} \& S-CNN $+$ SLFN ensemble) across $12$ Neovision2 Tower test sequences (in red).
  The top plot shows our WNMOTDA scores for individual classes (red symbols),
  while the vertical coloured bands indicate the approximate range of individual class scores obtained by the competing teams.
  The bottom plot shows the WNMOTDA detection scores,
  which are obtained by treating all objects (Car, Person, Cyclist) as a single class. 
  Our approach achieves state-of-the-art performance for Car and Cyclist, and comparable performance for Person. 
  The NMOTDA score for Person is reduced in cases when a single SEF tracks a group of people walking together, see text for details.  }
\end{figure}

Table~\ref{tab:results_totals} compares the total numbers of detections, false negatives, and true positives
with the total numbers of ground truth objects in our Tower test set of $12$ videos.
This indicates, for instance, that when considering all objects classes together, the total Recall is $\sim41\%$,
while the number of false positives per frame is $\sim 2.09$.
In Figure~\ref{fig:metrics_compare} we compare WNMOTDA with data points that we have extracted from the figures in~\cite{katsuri2014neuromorphicvision}.
Here Teams A, B and C rely on Neuromorphic Vision algorithms, whereas those denoted as Baseline are the results of a computer vision algorithm.
Our system is competitive with the state-of-the-art~\cite{khosla2014neuromorphic} (Team A) in terms of the detection score (Average WNMOTDA),
which demonstrates the efficacy of our track everything approach.
We also achieve the top scores for Cars and Cyclists, although it should be noted that this is on a reduced $12$ video test set.

\begin{table}[!htbp]
\begin{centering}
  \caption{\label{tab:results_totals} {Tower test set results across all 10452 frames: ground truths, detections, false negatives, false positives.}
  }
\par\end{centering}
\centering{}%
\begin{tabular}{l*{3}{c}r}
                        & $\sum_{t} \text{GT}_{t}$ & $\sum_{t} \text{Det}_{t}$ &  $\sum_{t} \text{FN}_{t}$ & $\sum_{t} \text{FP}_{t}$ \\
\hline
\hline
All                     & 82139   & 55271   & 48746    & 21878 \\
\hline
Car                     & 10452   & 11499   & 923      & 1970 \\
\hline
Person                  & 61699   & 35730   & 44558    & 18589 \\
\hline
Cyclist                 & 9988    & 8085    & 4294     & 2391 \\
\hline
\end{tabular}
\end{table}

\subsection{Discussion}

\subsubsection*{Prior knowledge}
We have shown that accurate online object recognition can be implemented
by using a general-purpose multi-object tracking system that is able to detect and track all salient objects.
For this to work, the use of object specific knowledge should be avoided.
We have identified in Table~\ref{tab:prior_knowledge} five sources of domain-specific prior knowledge used by {CACTuS-FL}:
the size of convolutional filters, the SEF shape size, the total number of SEFs, the scale of the second order moment ellipse
used to define bounding boxes, and the image patch size.
However, none of these parameters were tuned for specific object classes,
and therefore do not constitute object specific prior knowledge.

Team A~\cite{khosla2014neuromorphic} achieved state-of-the-art performance using an approach similar to ours, where salient objects are detected and prior knowledge is mostly embedded into the object classifier.  The saliency mechanism consists of fusing multiple saliency channels that are created from several individual feature response maps. However, prior knowledge is embedded into some of these saliency channels using the Targeted Contrast Enhancement (TCE) algorithm to create feature response maps that allow them to ``easily detect objects with [specified] colors, \ldots\ e.g. finding all red cars on the road.''   Another point of difference is that Team A do not perform tracking, only detection and classification.  Instead they embed motion processing as another saliency channel, which detects pixels that appear to be moving in comparison to a (stationary or registered) background scene.

The primary difference between our approach and traditional tracking-by-detection approaches is that prior knowledge of the objects of interest is removed from detection and tracking, and only used for recognition.  

\subsubsection*{Advantages}

The advantage is that all objects are tracked and `explained away', including sources of clutter.  This handling of distracting and occluding clutter improves tracking robustness~\cite{wong2014modeldrift}. For instance, when a person (target) walks behind a lamppost (clutter), the SEF tracking the lamppost learns that it is not moving and the competitive attentional mechanisms in CACTuS-FL allow the SEF tracking the person to ignore the observations from the lamppost. 

\subsubsection*{Limitations}

One limitation in our current approach is that the tracking system does not know what the extent of a single object is; it simply associates a consistent set of observations (in shape, position and velocity) with a single SEF.  For example, people walking together in a group (thus having the same position and velocity) can be efficiently described in the state-space of a single SEF, and thus be considered a single object.  This occurs in the Neovision2 Tower test data set video 023. Here a single SEF tracks a crowd of people and the classifier labels the track a `Person'.  However, the bounding box of the crowd is larger than the ground-truth box of any individual person, thus failing the spatial overlap requirement $d_{t,j}  > T_d$ from Eqn.~(\ref{eq:overlap}).  This results in both one false positive for the SEF tracking the crowd and many false negatives for the individual people within the crowd, and thus a poor NMOTA score of -0.26 for the video (see supplemental material). This video is a key contributor to the low WNMOTA for the Person class in Figure~\ref{fig:metrics_compare}. Furthermore, without this video the overall Average WNMOTA score would be $0.17$ rather than its present value of $0.14$. 

\subsubsection*{Integrating what and where}

In our architecture low level processing is performed with a common set of convolutional filters (see Figure~\ref{fig:crbm_filters}),
resulting in a shared set of features for the separate \emph{what} and \emph{where} processing streams.
The \emph{what} processing stream is performed by the S-CNN, while the \emph{where} processing stream is performed using CACTuS-FL.
By parameterizing elements of the CACTuS-FL state information, it is possible to efficiently re-integrate the \emph{what} and \emph{where} processing stream, using the SLFN ensemble.
The benefit of the integration is a gain in Average WNMOTDA of $0.4$ as shown in Figure~\ref{fig:metrics_2nd}.
This improvement in recognition performance may provide insight into the function of neurons
that integrate both the \emph{what} and \emph{where} processing streams in the primate visual cortex~\cite{rao1997integration}.
Knowing \emph{where} an object is (tracking) may help recognise \emph{what} an object is (classification).

\section{Conclusion}
\label{sec:conclusion}

We have presented a system for online object recognition that can autonomously locate and recognize multiple types of objects
using biologically inspired \emph{what} and \emph{where} processing streams.
Our overall approach may be characterized as a shift of the use of object-specific prior knowledge out of the \emph{where} stream and into the \emph{what} stream.
This enables the \emph{where} stream, which is implemented as a general purpose multi-object tracking algorithm,
to locate every salient object in the scene, including sources of occluding or distracting clutter.
Online recognition of localized objects is then handled by re-integration of the \emph{what} and \emph{where} processing streams.
This takes the form of a SLFN ensemble that combines object-tracking state information 
with class label estimate information from the S-CNN to provide robust object recognition outputs, the performance of which is comparable to the state-of-the-art.

\section{Acknowledgement}

This work was completed in part with the support provided by the Defence Science and Technology Group, Australian Department of Defence, Commonwealth of Australia.


%





\ifCLASSOPTIONcaptionsoff
  \newpage
\fi



\bibliographystyle{IEEEtran}
\bibliography{IEEEabrv,bib/prior_knowledge_refs}

\begin{IEEEbiography}[{\includegraphics[width=1in,height=1.25in,clip,keepaspectratio]{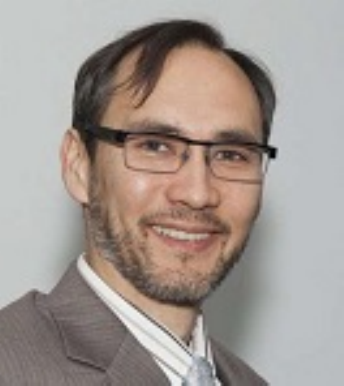}}]{Sebastien Wong}
(Senior Member, IEEE) joined the Defence Science and Technology (DST) Australia in 1999 where he currently holds the position of Science Team Leader Image Processing.  He is also the Director of Machine Learning at Consilium Technology, and an Adjunct Associate Professor at the University of South Australia. Sebastien was awarded the Chief of Air force Gold Commendation in 2008 for his work on missile approach warning algorithms, and a DSTO Achievement Award in 2012 for his work on Hostile Fire Indication algorithms. Sebastien holds a bachelor's degree in Computer Systems Engineering (with honours) from Curtin University, a master's degree in Electronic Systems Engineering and a PhD in Computer Science, both from the University of South Australia, as well as graduate diploma in Scientific Leadership from the University of Melbourne. Sebastien has published over 20 peer-reviewed papers. His research interests include machine learning algorithms and parallel processing architectures for autonomous vision systems. He is IEEE senior member and a past chair for the IEEE South Australia Section Computer Society Chapter.
\end{IEEEbiography}


\begin{IEEEbiography}[{\includegraphics[width=1in,height=1.25in,clip,keepaspectratio]{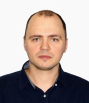}}]{Victor Stamatescu}
(Member, IEEE) received a Bachelor of Science (Hons), majoring in Physics, and a PhD in Astrophysics, both from the University of Adelaide in 2004 and 2010, respectively. Between 2010 and 2013 he was a postdoctoral research scientist at the Institute of High Energy Physics (IFAE) in Barcelona, Spain. From 2013 to 2014 he was a research fellow in the High Energy Astrophysics group at the University of Adelaide. Since 2014 he has been a research fellow in the School of Information Technology and Mathematical Sciences at the University of South Australia. Victor has published 79 peer-reviewed journal and conference papers. His current research interests are in the areas of visual tracking, image classification and machine learning.
\end{IEEEbiography}


\begin{IEEEbiography}[{\includegraphics[width=1in,height=1.25in,clip,keepaspectratio]{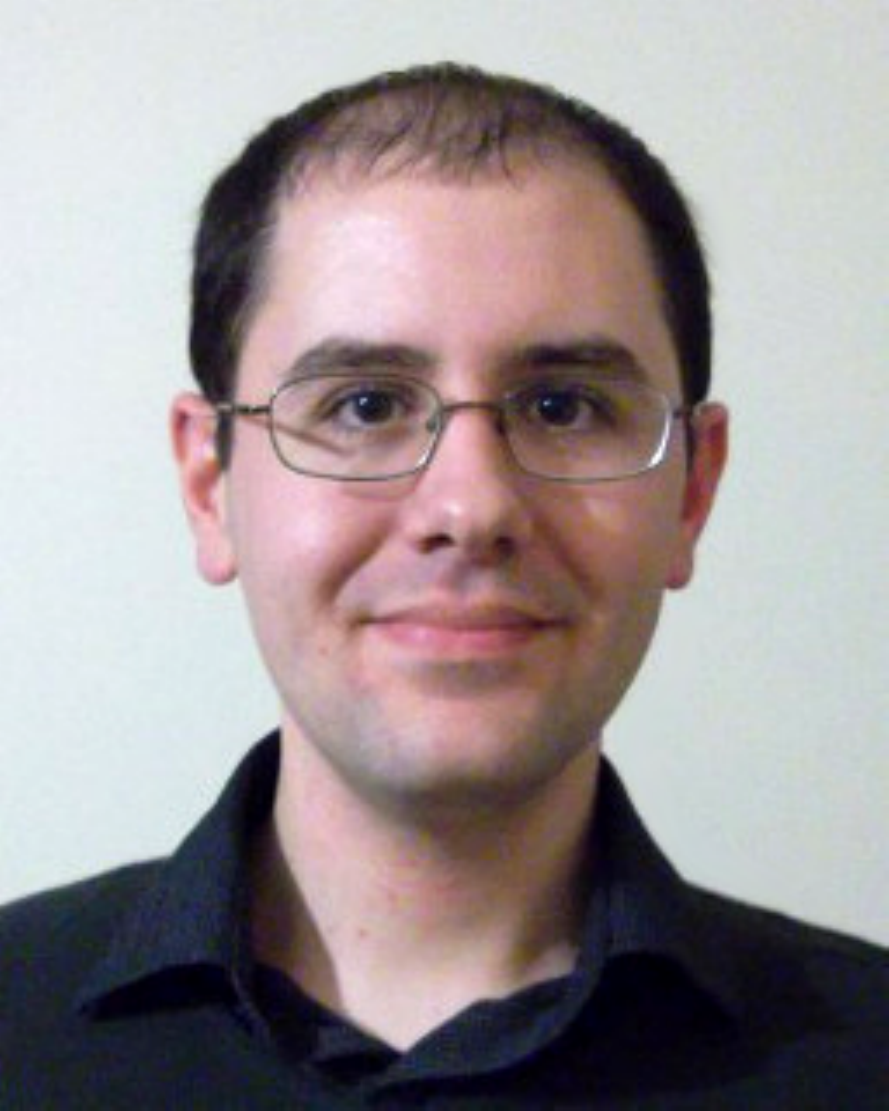}}]{Adam Gatt}
(Member, IEEE) is a software developer with the Australian Department of Defence. His research interests involve visual tracking, feature learning and tracker evaluation. Adam was awarded a PhD in Computer Science from the University of South Australia in 2013. He received a bachelor's degree in Information Technology with first class honours in 2008, and was awarded the Joyner Scholarship and UniSA Vice Chancellor and President's Scholarship in 2009. He is a member of the IEEE Computer Society and is a past chair for the South Australia Chapter.
\end{IEEEbiography}


\begin{IEEEbiography}[{\includegraphics[width=1in,height=1.25in,clip,keepaspectratio]{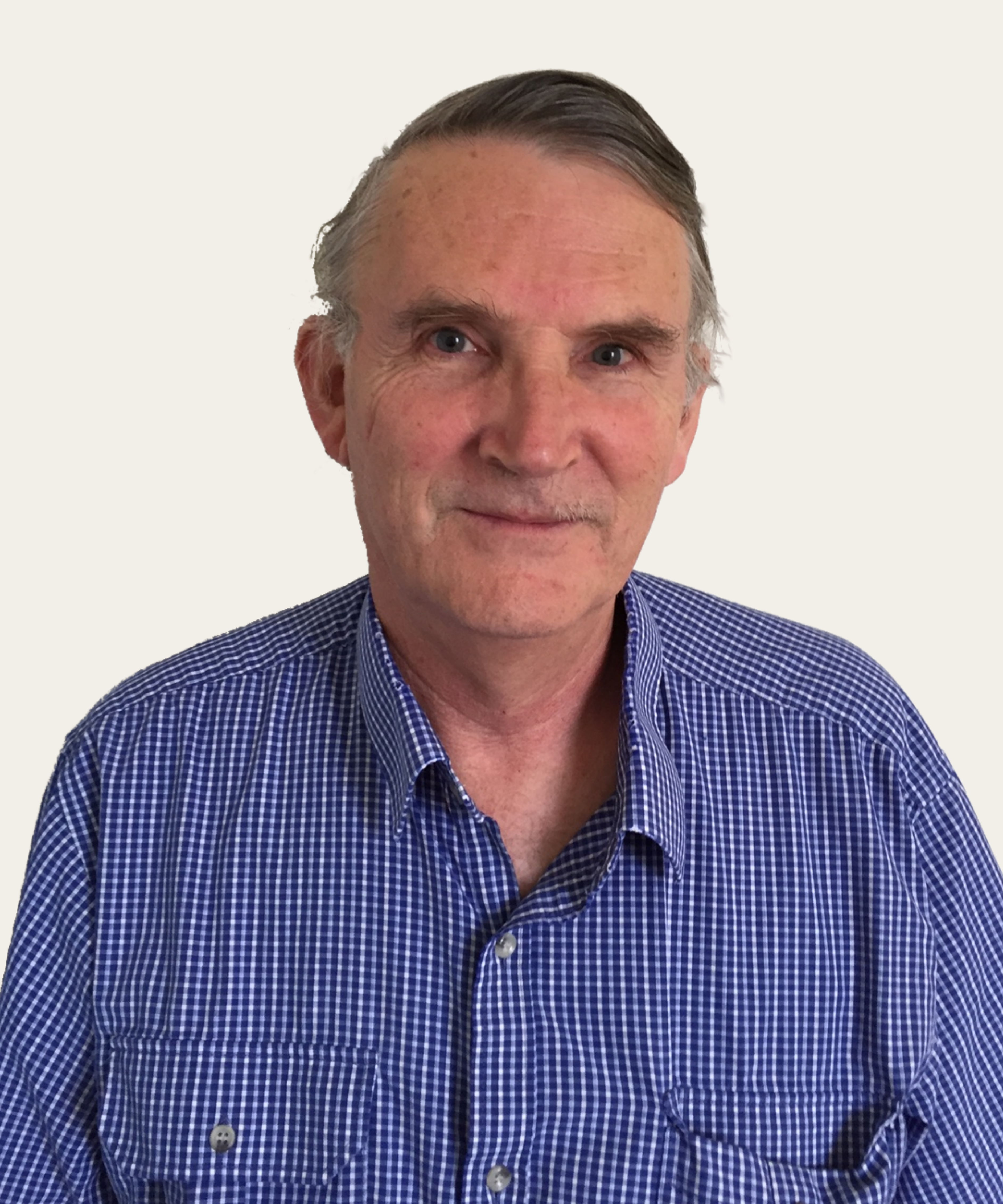}}]{David Kearney}
is Associate Professor of Computer Science at the University of South Australia. David's research has focused on high performance parallel computing using reconfigurable hardware based on field programmable gate arrays, leading to over 90 refereed publications. His research outputs include the Hardware Join Java language and the ReconfigME operating system for reconfigurable computing. He has interests in applications related to high speed parallel image processing, simulation and tracking. David has in recent years taken an interest in forms of parallel computing inspired by biology including membrane computing and neural networks.
\end{IEEEbiography}


\begin{IEEEbiography}[{\includegraphics[width=1in,height=1.25in,clip,keepaspectratio]{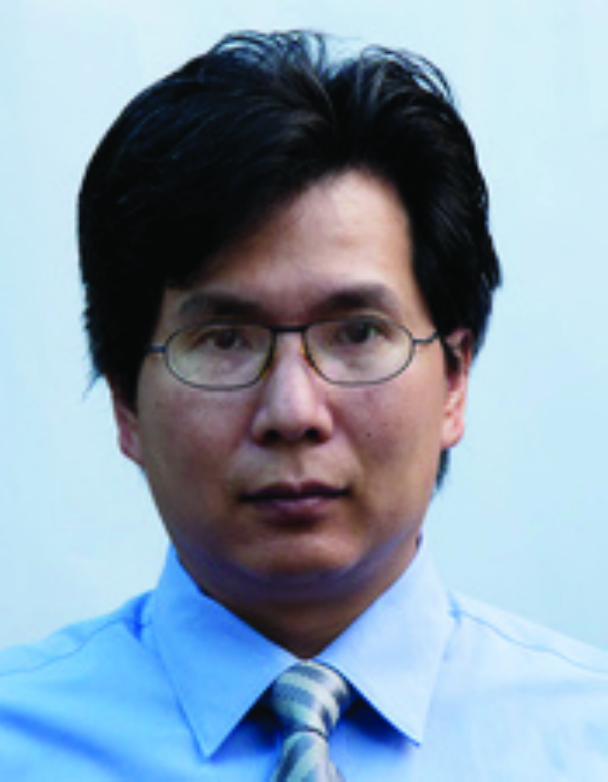}}]{Ivan Lee}
(Senior Member, IEEE) received BEng, MCom, MER, and PhD degrees from the University of Sydney, Australia. He was a software development engineer at Cisco Systems, a software engineer at Remotek Corporation, and an Assistant Professor at Ryerson University. Since 2008, he has been a Senior Lecturer at the University of South Australia. His research interests include multimedia systems, medical imaging, data analytics, and computational economics.
\end{IEEEbiography}


\begin{IEEEbiography}[{\includegraphics[width=1in,height=1.25in,clip,keepaspectratio]{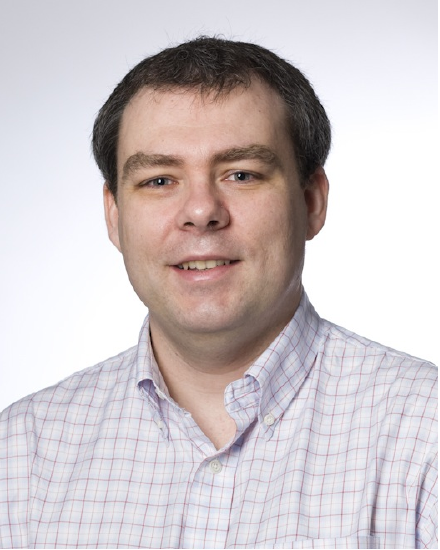}}]{Mark D. McDonnell}
(Senior Member, IEEE) received the B.E. degree in electronic engineering, the B.Sc. degree (with first class honors) in applied mathematics, and the Ph.D. degree in electronic engineering from The University of Adelaide, Australia, in 1998, 2001, and 2006, respectively. He is currently Associate Professor, and Principal Investigator of the Computational Learning Systems Laboratory at the University of South Australia, which he joined in 2007. He has published over 100 papers, including several review articles, and a book on stochastic resonance, published by Cambridge University Press. Prof. McDonnell has served as a Guest Editor for Proceedings of the IEEE and Frontiers in Computational Neuroscience. His research interests lie in the intersection between machine learning and neurobiological learning, with a specific focus on the influence of random noise on learning. His contributions to this area have been recognized by the award of an Australian Research Fellowship from the Australian Research Council in 2010, and a South Australian Tall Poppy Award for Science in 2008. He has served as Vice President and Secretary of the IEEE South Australia Section Joint Communications and Signal Processing Chapter.
\end{IEEEbiography}

\end{document}


\maketitle

\emph{Neovision2} Tower test set object recognition results obtained using \emph{CACTuS-FL} \& Shallow CNN $+$ SLFN ensemble. Listed are the numbers of Ground Truths (GT), Detections (Det), False Negatives (FN), False Positives (FP), number of image frames (Frames) and Normalized Multiple Object Thresholded Detection Accuracy (NMOTDA), for each video.

\begin{table}[!htbp]
\begin{centering}
  \caption{\label{tab:results_person} {\emph{Neovision2} Tower test set results obtained for All classes.}}
\par\end{centering}
\centering{}%
\begin{tabular}{l*{5}{c}r}
Test Sequence           & All GT & Det & FN & FP & Frames & NMODTA\\
\hline
\hline
002                     & 8499  &  5222 & 4924  & 1647 & 871 & 0.226850 \\
\hline
003                     & 9761  &  5414 & 6892  & 2545 & 871 & 0.033193 \\
\hline
009                     & 7304  &  5551 & 3352  & 1599 & 871 & 0.322152 \\
\hline
010                     & 5161  &  3580 & 2610  & 1029 & 871 & 0.294904 \\
\hline
012                     & 5680  &  3332 & 2803  &  455 & 871 & 0.426408 \\
\hline
013                     & 9779  &  5594 & 7243  & 3058 & 871 & -0.053380 \\
\hline
017                     & 3827  &  5621 &  666  & 2460 & 871 & 0.183172 \\
\hline
018                     & 7909  &  3251 & 5468  &  810 & 871 & 0.206221 \\
\hline
019                     & 7420  &  5467 & 3428  & 1475 & 871 & 0.339218 \\
\hline
021                     & 4925  &  3142 & 3177  & 1394 & 871 & 0.071878 \\
\hline
022                     & 5847  &  4349 & 3755  & 2257 & 871 & -0.028220 \\
\hline
023                     & 6027  &  4748 & 4428  & 3149 & 871 & -0.257176 \\
\hline
\end{tabular}
\end{table}

\begin{table}[!htbp]
\begin{centering}
  \caption{\label{tab:results_person} {\emph{Neovision2} Tower test set results obtained for the Person class.}}
\par\end{centering}
\centering{}%
\begin{tabular}{l*{5}{c}r}
Test Sequence           & Person GT & Det & FN & FP & Frames & NMODTA\\
\hline
\hline
002                     & 6556  & 3455  & 4677  & 1576 & 871 & 0.046217 \\
\hline
003                     & 7643  & 3458  & 6252  & 2067 & 871 & -0.088447 \\
\hline
009                     & 4890  & 3291  & 2945  & 1346 & 871 & 0.122495 \\
\hline
010                     & 3138  & 2002  & 2049  &  913 & 871 & 0.056087 \\
\hline
012                     & 4190  & 1840  & 2813  &  463 & 871 & 0.218138 \\
\hline
013                     & 8257  & 4232  & 6992  & 2967 & 871 & -0.206128 \\
\hline
017                     & 2766  & 3985  & 541   & 1760 & 871 & 0.168113 \\
\hline
018                     & 6506  & 2481  & 4786  &  761 & 871 & 0.147402 \\
\hline
019                     & 5541  & 3875  & 2952  & 1286 & 871 & 0.235156 \\
\hline
021                     & 3738  & 1883  & 2986  & 1131 & 871 & -0.101391 \\
\hline
022                     & 4356  & 2861  & 3594  & 2099 & 871 & -0.306933 \\
\hline
023                     & 4118  & 2367  & 3971  & 2220 & 871 & -0.503400 \\
\hline
\end{tabular}
\end{table}

\begin{table}[!htbp]
\begin{centering}
  \caption{\label{tab:results_cyclist} {\emph{Neovision2} Tower test set results obtained for the Cyclist class.}}
\par\end{centering}
\centering{}%
\begin{tabular}{l*{5}{c}r}
Test Sequence           & Cyclist GT & Det & FN & FP & Frames & NMODTA\\
\hline
\hline
002                     &  1072 &  938  & 406  & 272  & 871 & 0.367537 \\
\hline
003                     &  1247 &  659  & 727  & 139  & 871 & 0.305533 \\
\hline
009                     &  1543 &  1402 & 543  & 402  & 871 & 0.387557 \\
\hline
010                     &  1152 &  728  & 589  & 165  & 871 & 0.345486 \\
\hline
012                     &  619  &  626  & 137  & 144  & 871 & 0.546042 \\
\hline
013                     &  651  &  499  & 258  & 106  & 871 & 0.440860 \\
\hline
017                     &  190  &  78   & 142  & 30   & 871 & 0.094737 \\
\hline
018                     &  532  &  586  & 136  & 190  & 871 & 0.387218 \\
\hline
019                     &  1008 &  790  & 467  & 249  & 871 & 0.289683 \\
\hline
021                     &  316  &  416  & 201  & 301  & 871 & -0.588608 \\
\hline
022                     &  620  &  651  & 238  & 269  & 871 & 0.182258 \\
\hline
023                     &  1038 &  712  & 450  & 124  & 871 & 0.447013 \\
\hline
\end{tabular}
\end{table}

\begin{table}[!htbp]
\begin{centering}
  \caption{\label{tab:results_car} {\emph{Neovision2} Tower test set results obtained for the Car class.}}
\par\end{centering}
\centering{}%
\begin{tabular}{l*{5}{c}r}
Test Sequence           & Car GT & Det & FN & FP & Frames & NMODTA\\
\hline
\hline
002                     &  871  & 830   &  41  &  0 & 871 & 0.952928 \\
\hline
003                     &  871  & 1298  &  32  & 459 & 871 & 0.436280 \\
\hline
009                     &  871  & 865   &  10  &  4 & 871 & 0.983927 \\
\hline
010                     &  871  & 851   &  24  & 4 & 871 & 0.967853 \\
\hline
012                     &  871  & 867   &  4  & 0 & 871 & 0.995408 \\
\hline
013                     &  871  & 865   &  6  & 0 & 871 & 0.993111 \\
\hline
017                     &  871  & 1560  &  6  & 695 & 871 & 0.195178 \\
\hline
018                     &  871  & 199   &  672  &  0 & 871 & 0.228473 \\
\hline
019                     &  871  & 802   &  69  &  0 & 871 & 0.920781 \\
\hline
021                     &  871  & 856   &  15  &  0 & 871 & 0.982778 \\
\hline
022                     &  871  & 837   &  34  &  0  & 871 & 0.960964 \\
\hline
023                     &  871  & 1669  &  10  & 808 & 871 & 0.060850 \\
\hline
\end{tabular}
\end{table}